%% file: main.tex
\documentclass{article}

\usepackage{graphicx}
\usepackage{booktabs}
\usepackage[table]{xcolor}
\usepackage{subfig}
\usepackage{multirow}
\usepackage{amsmath}
\usepackage{amssymb}
\usepackage{pifont}
\usepackage{CJKutf8}
\usepackage{caption}

\usepackage[preprint]{neurips_2025}

\usepackage[utf8]{inputenc} 
\usepackage[T1]{fontenc}    
\usepackage{hyperref}       
\usepackage{url}            
\usepackage{booktabs}       
\usepackage{amsfonts}       
\usepackage{nicefrac}       
\usepackage{microtype}      
\usepackage{xcolor}         

\usepackage[utf8]{inputenc}
\usepackage[T1]{fontenc}
\usepackage{amsmath, amssymb}
\usepackage{graphicx}
\usepackage[dvipsnames, table]{xcolor}
\usepackage{geometry}
\usepackage{booktabs}
\usepackage{tabularx}
\usepackage{enumitem}
\usepackage{setspace}
\usepackage[export]{adjustbox}
\usepackage{etoc}
\usepackage{changepage}
\usepackage{makecell}

\title{VIR-Bench: Evaluating Geospatial and Temporal Understanding of MLLMs via Travel Video Itinerary Reconstruction}

\author{%
  Hao~Wang\thanks{Equal contribution. Correspondence to Hao Wang (\texttt{conan1024hao@akane.waseda.jp})~.}~~$^{1}$ \quad Eiki Murata\footnotemark[1]~~$^{2,3}$ \quad Lingfang Zhang$^{1}$ \vspace{-20mm}\AND
  Ayako Sato$^{2}$ \quad So Fukuda$^{1}$ \quad Ziqi Yin$^{1}$ \quad Wentao Hu$^{1}$ \vspace{-20mm}\AND
  Keisuke Nakao$^{1}$ \quad Yusuke Nakamura$^{1}$ \quad Sebastian Zwirner$^{1}$ \quad Yi-Chia Chen$^{1}$ \vspace{-20mm}\AND
  Hiroyuki Otomo$^{2}$ \quad  Hiroki Ouchi$^{4,2}$ \quad Daisuke Kawahara$^{1}$ \vspace{5mm} \\
  \textsuperscript{1}Waseda University \quad
  \textsuperscript{2}CyberAgent, Inc. \quad
  \textsuperscript{3}AI Shift, Inc.  \\
  \textsuperscript{4}Nara Institute of Science and Technology \vspace{5mm} \\
  \small{\url{https://github.com/nlp-waseda/VIR-Bench}}
}

\begin{document}

\maketitle

\begin{abstract}
Recent advances in multimodal large language models (MLLMs) have significantly enhanced video understanding capabilities, opening new possibilities for practical applications.
Yet current video benchmarks focus largely on indoor scenes or short-range outdoor activities, leaving the challenges associated with long-distance travel largely unexplored.
Mastering extended geospatial-temporal trajectories is critical for next-generation MLLMs, underpinning real-world tasks such as embodied-AI planning and navigation.
To bridge this gap, we present \textbf{VIR-Bench}, a novel benchmark consisting of 200 travel videos that frames itinerary reconstruction as a challenging task designed to evaluate and push forward MLLMs' geospatial-temporal intelligence.
Experimental results reveal that state-of-the-art MLLMs, including proprietary ones, struggle to achieve high scores, underscoring the difficulty of handling videos that span extended spatial and temporal scales.
Moreover, we conduct an in-depth case study in which we develop a prototype travel-planning agent that leverages the insights gained from VIR-Bench.
The agent’s markedly improved itinerary recommendations verify that our evaluation protocol not only benchmarks models effectively but also translates into concrete performance gains in user-facing applications.
\end{abstract}

\section{Introduction}
\begin{figure*}[t]
  \centering
  \includegraphics[width=\linewidth]{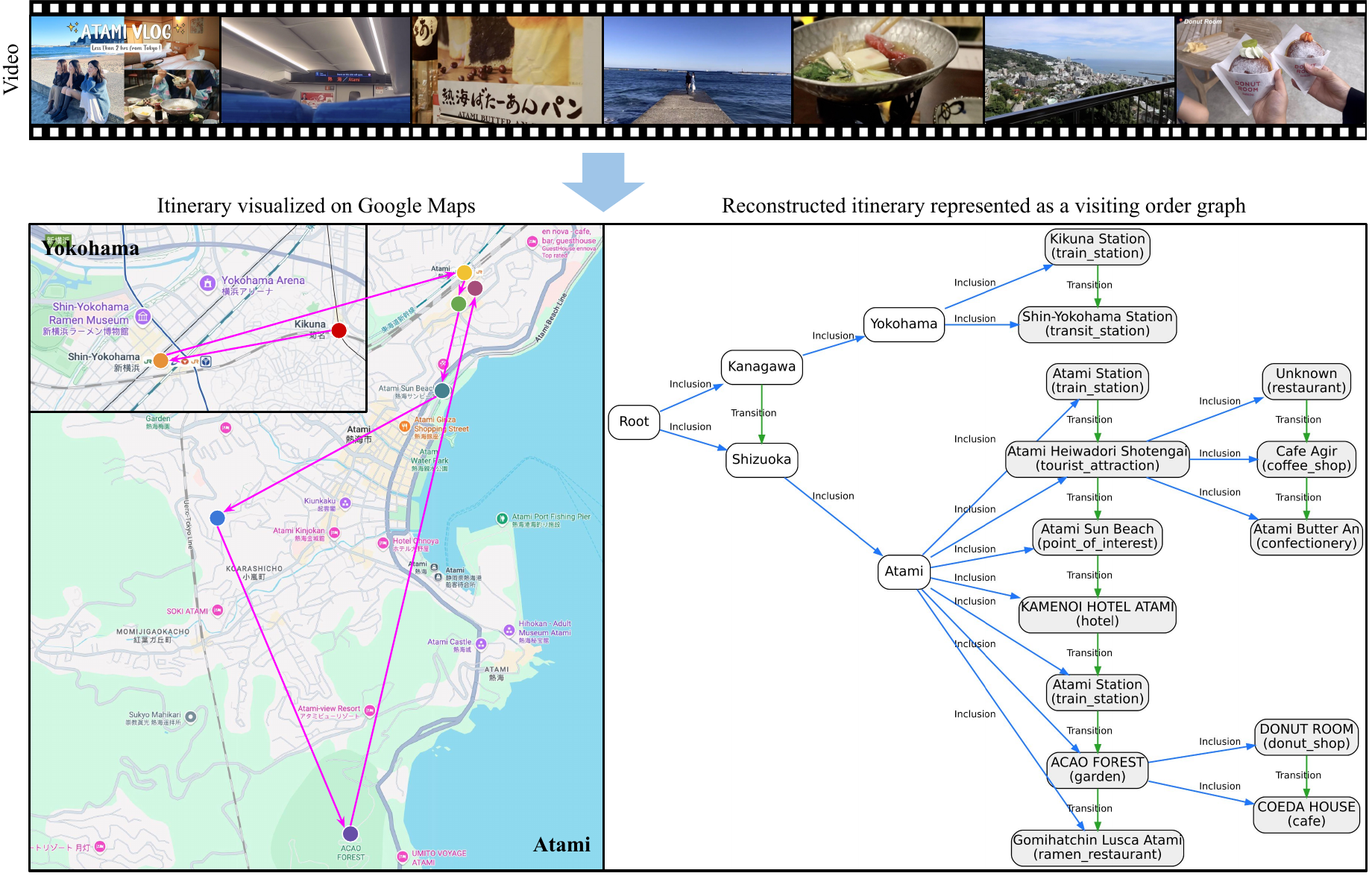}
  \caption{Overview of VIR-Bench. Given an input travel video (\textbf{Top}), we reconstruct a visiting order graph (\textbf{Right}) whose nodes are visited locations (prefectures, cities, and POIs) and whose edges capture both temporal transitions and geographic containment among the locations. The itinerary visualization (\textbf{Left}) omits the second stop at Atami Station for visual clarity. The video frames are adopted from \url{https://www.youtube.com/watch?v=6aJ4CZfn9c8}.}
  \label{fig:main-figure}
\end{figure*}
Recent advances in multimodal large language models (MLLMs)~\cite{liu2023visualinstructiontuning,li2024llavanextinterleavetacklingmultiimagevideo,Lin_2024_CVPR,openai2024gpt4technicalreport} have improved remarkable capabilities in video understanding.
Lately, research attention has shifted toward evaluating the spatial and temporal reasoning abilities of MLLMs, prompting the proposal of new benchmarks~\cite{grauman2022ego4dworld3000hours,chandrasegaran2024hourvideo,jia2024sceneversescaling3dvisionlanguage,yang2025thinking,lin2025ost}.
However, existing benchmarks primarily focus on micro-scale scenarios, such as indoor scenes or short-range outdoor activities, leaving macro-scale geospatial scenarios, namely, long-distance travel activities involving multi-day footage across multiple cities, largely unexplored.
We argue that long-horizon geospatial-temporal reasoning is essential for next-generation MLLMs, as numerous real-world applications, such as embodied AI planning, navigation, and autonomous driving, heavily rely on these capabilities.

To address this gap, we introduce VIR-Bench, a benchmark to evaluate long-range geospatial-temporal understanding via itinerary reconstruction from travel vlog videos.
The core output is a directed visiting order graph~\cite{yamamoto-etal-2025-graph}: nodes represent locations at three granularities (prefecture, city, and point of interest (POI)) and edges represent two relations, inclusion for spatial hierarchy and transition for temporal adjacency.
By decomposing the task into two sub-tasks: (1) node prediction, identifying all locations visited; and (2) edge prediction, inferring geographic inclusion relations and temporal transition relations among visited locations, VIR-Bench enables separate evaluation of geospatial and temporal intelligence.
Because the footage is mostly egocentric or selfie-style, models must construct a holistic understanding from partial views, which further stresses geospatial–temporal reasoning.
The dataset comprises 200 travel vlogs filmed across Japan, a major inbound tourism destination, each accompanied by a manually annotated and double-reviewed visiting order graph.

Through extensive experiments on state-of-the-art open-weight and proprietary MLLMs, we observe persistent challenge in geospatial and temporal understanding.
Particularly, open-weight models suffer from insufficient geographic knowledge and limited capability for long-context reasoning.
Although proprietary models achieve better performance, they still struggle on POI node prediction and transition edge prediction, which remain major bottlenecks.
Ablations further reveal that more visual context (frames), greater reasoning effort, and access to audio each provide consistent gains.
Collectively, these findings highlight key obstacles that need to be addressed to advance geospatial-temporal applications in real-world settings.

In addition, we develop a prototype travel-planning agent that generates travel plans directly from videos and their corresponding visiting order graphs.
Results from crowdsourcing and automatic evaluations indicate that the itinerary, represented by the POI list, is essential for producing logically sound and feasible plans, underscoring the importance of itinerary reconstruction.
Meanwhile, the video provides rich, nuanced context that enhances the attractiveness of a travel plan.
A setting that uses both the itinerary and the video leverages these complementary strengths, highlighting a promising approach to generating high-quality travel plans from multimedia sources.
These findings validate that our benchmark not only effectively evaluates models but also pushes forward practical user-facing applications.
\section{Related Work}
\subsection{Video Benchmarks}
    With the rapid advancement of MLLMs, recent video understanding benchmarks have increasingly prioritized evaluating models' spatial and temporal reasoning capabilities.
    For instance, Ego4D~\cite{grauman2022ego4dworld3000hours} facilitates the evaluation of models' comprehension of past and future events by utilizing curated temporal data, while HourVideo~\cite{chandrasegaran2024hourvideo} examines the performance of models in understanding extended-duration video content.
    VSI-Bench~\cite{yang2025thinking} assesses a model's ability to infer 3D scene layouts from 2D video inputs, while OST-Bench~\cite{lin2025ost} evaluates spatial-temporal understanding by requiring models to explore and interpret information within a 3D space.
    CityGuessr~\cite{kulkarni2024cityguessrcitylevelvideogeolocalization} introduces a video-based benchmark for assessing geo-localization using driving videos, while UrbanVideo-Bench~\cite{zhao2025urbanvideobenchbenchmarkingvisionlanguagemodels} targets the embodied cognitive abilities of MLLMs within urban 3D environments using drone-collected footage.
    Nevertheless, most existing benchmarks primarily feature indoor scenarios or short-distance outdoor movements, lacking extensive long-distance traversal, such as inter-city journeys.
    Consequently, these benchmarks are insufficient for thoroughly evaluating the geospatial-temporal intelligence of MLLMs.
    In contrast, VIR-Bench specifically addresses this gap by comprehensively assessing video understanding capabilities across extended spatial (e.g., from Tokyo to Osaka, Hokkaido to Kyoto) and temporal (spanning multiple days) scales.
\subsection{Itinerary Extraction}
    Researchers in natural language processing have substantially studied the task of extracting travel trajectories from text~\cite{10.1145/1869890.1869894,kaushik-etal-2017-making,ijgi10100710,yamamoto-etal-2025-graph}.
    A representative study by \citet{yamamoto-etal-2025-graph} introduces a visiting order graph designed to capture relationships among visited locations and provides a benchmark dataset for training and evaluating itinerary extraction models.
    In multimodal settings, \citet{PANG2011352} proposes a framework aimed at summarizing travelogues by integrating text and images from blogs.
    More recently, \citet{DBLP:conf/rectour/Rosa24} leverages MLLMs to perform structured entity extraction from travel videos, while \citet{zhuang2024vloggermakedreamvlog} tackles the inverse problem by generating vlogs with diverse travel scenes.
    This study advances this line of work by providing, to our knowledge, the first systematic investigation into extracting and reconstructing itineraries directly from videos, establishing a new benchmark for video-centric geospatial and temporal understanding.
\subsection{Itinerary Generation}

    The complexity of manual trip planning has driven research into automated itinerary generation.
    Initial approaches were often based on optimization problems like the Tourist Trip Design Problem and classic machine learning~\cite{Gavalas2014, 36046891aa77462298f21dacee3279a6, 8731399, Carrillo2023SmartTurECO}.
    More recently, Large Language Models (LLMs) have enabled more sophisticated and flexible frameworks~\cite{chen2024travelagentaiassistantpersonalized, xie2024travelplanner}.
    Current research trends include developing novel reasoning paradigms~\cite{gui2025hypertree}, creating hybrid systems that combine LLMs with classical planners~\cite{delarosa2024trippaltravelplanningguarantees}, and enhancing personalization through user model integration and interactive feedback~\cite{singh-etal-2024-personal, chen2024travelagentaiassistantpersonalized, OTAKI2025126294}.

    To rigorously evaluate these methods, various benchmarks have been developed.
    Notable studies include real-world planning~\cite{xie2024travelplanner}, fine-grained spatio-temporal planning~\cite{Chaudhuri2025TripCraft}, and assessing personalization~\cite{singh-etal-2024-personal}.

    Existing benchmarks and generation methods primarily use text data such as user preferences and travel logs as input.
    In contrast, this research uses travel videos as input, aiming to reconstruct the itinerary based on their content.
\section{Dataset Construction}
VIR-Bench comprises 200 travel videos filmed across Japan, each paired with a corresponding visiting order graph that captures the itinerary depicted in the video.
In this section, we present the construction process of VIR-Bench.
We begin by defining the visiting order graph, followed by a detailed description of our data annotation procedure.
    \begin{figure}[t]
    \centering
    \includegraphics[width=0.75\linewidth]{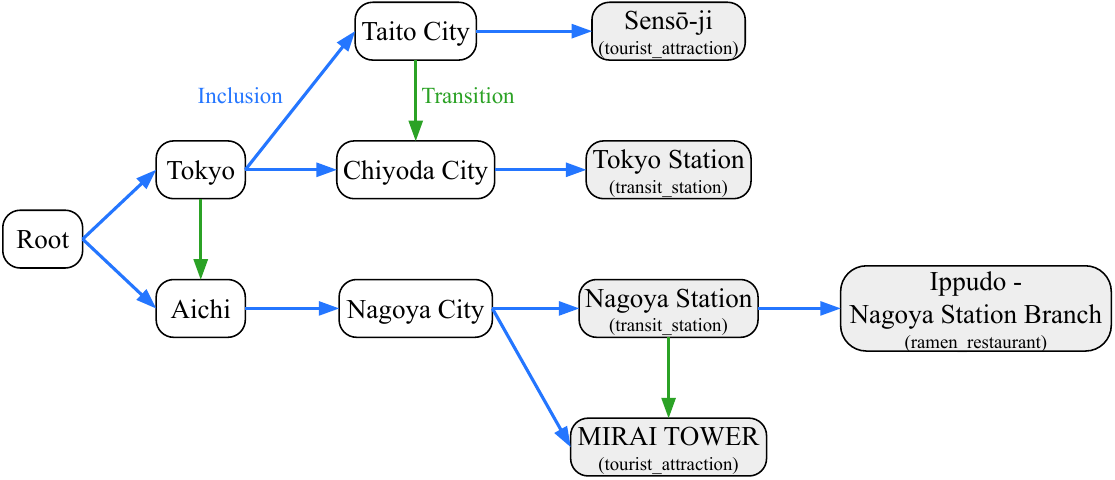}
    \caption{Example of a visiting order graph. \textbf{Inclusion edges} represent containment relationships, flowing from a larger geographical area to a smaller one. \textbf{Transition edges} indicate chronological movement between distinct locations at the same hierarchical level.}
    \label{fig:data-example}
    \end{figure}
\subsection{Visiting Order Graph}
    We adopt and refine the definition of a visiting order graph introduced by \citet{yamamoto-etal-2025-graph}.
    A simplified example is shown in Figure~\ref{fig:data-example}.
    
    A visiting order graph is a hierarchical directed graph with four node types:
    \begin{itemize}
        \item \textbf{Root node}: the starting node of the graph.
        \item \textbf{Prefecture node}: the highest-level administrative division (e.g., Tokyo, Osaka, Aichi).
        \item \textbf{City node}: a municipality within a prefecture, including Tokyo's special wards, cities, towns, and villages.
        \item \textbf{POI node}: a specific named location (point of interest), such as landmarks, tourist attractions, stations, restaurants, cafes, stores, parks, or museums.
    \end{itemize}
    The graph includes two edge types:
    \begin{itemize}
        \item \textbf{Inclusion edge}: a directed edge representing containment of one location within another. This edge flows from the larger geographical area to the smaller one.
        \begin{itemize}
            \item Prefecture $\rightarrow$ City (e.g., \textit{Aichi} $\rightarrow$ \textit{Nagoya}).
            \item City $\rightarrow$ POI (e.g., \textit{Nagoya} $\rightarrow$ \textit{Nagoya Station}).
            \item POI $\rightarrow$ sub-POI (e.g.,\\\textit{Nagoya Station} $\rightarrow$ \textit{Ippudo - Nagoya Station Branch}).
        \end{itemize}
        \item \textbf{Transition edge}: a directed edge representing movement between two distinct locations at the same hierarchical level; it indicates the chronological flow of travel.
        \begin{itemize}
            \item Between prefectures (e.g., \textit{Tokyo} $\rightarrow$ \textit{Aichi}).
            \item Between cities within the same prefecture\\(e.g., \textit{Taito City} $\rightarrow$ \textit{Chiyoda City}).
            \item Between POIs within the same city\\(e.g., \textit{Nagoya Station} $\rightarrow$ \textit{MIRAI TOWER}).
        \end{itemize}
    \end{itemize}
    
    To prevent cycles in the graph, we treat multiple visits to the same location as distinct nodes.
    Following \citet{yamamoto-etal-2025-graph}, we also introduce a special ``Overlap'' edge to handle POIs that are geographically overlapping but cannot be represented through inclusion edges.
\subsection{Data Annotation}
    Identifying locations visited in travel videos is similar to playing GeoGuessr\footnote{\url{https://www.geoguessr.com}}: annotators infer places from visual cues.
    We recruited 10 Japan-based annotators, each tasked with collecting 20 YouTube travel videos filmed in Japan.
    The videos could be narrated in English or Japanese.
    Annotators were asked to identify all visited POIs in each video.
    We define a ``visit'' as when the POI appears in the video and it is clear that the videographer visited the facility.
    For every POI, they recorded the start and end times within the video and provided the corresponding Google Maps URL.
    When a location could not be uniquely identified (e.g., a cafe shown without its name), they entered the placeholder \texttt{UNKNOWN} and recorded the POI category (e.g., \texttt{cat\_cafe}).
    
    After annotation, we retrieved detailed information of each POI including name, address and categories using Google Places API.
    Using the timestamped POIs, we then constructed a visiting order graph for each video (e.g., Figure~\ref{fig:data-example}).
    We also conducted a quality check of the generated graphs and corrected errors by rerunning the retrieval  step manually when POIs were incorrectly annotated or matched.
    This pipeline yielded VIR-Bench, a dataset of 200 travel videos (100 in English and 100 in Japanese) paired with their corresponding visiting order graphs.
    In total, 3,689 POIs were identified across 43 of Japan’s 47 prefectures.
    Detailed annotation guidelines and dataset statistics are provided in Appendix~\ref{benchmark-details}.


\section{Experiments}
\subsection{Task Definition}
    We aim to generate visiting order graphs directly from videos with MLLMs.
    However, our preliminary experiments revealed that this end-to-end approach is too difficult for current models.
    To address this, we decompose the task into two sub-tasks: node prediction and edge prediction.
    We describe each of these tasks in the following.
    \paragraph{Node Prediction}
    This task evaluates models’ geospatial understanding, akin to playing ``GeoGuessr’’.
    Given a video, MLLMs are asked to return all visited locations in three JSON lists (prefectures, cities, and POIs).
    For each POI, the model must also predict its category.
    \paragraph{Edge Prediction}
    Given a video and all visited locations (gold labels, shuffled), MLLMs are asked to predict all inclusion and transition edges that constitute the video’s visiting order graph.
    The output should be a JSON list of tuples formatted as \texttt{<source, target, edge\_type>}.
    Inclusion edge prediction evaluates models’ geospatial knowledge, while transition edge prediction assesses their temporal understanding.
    We omit overlap edges in this task due to their low frequency.
\subsection{Benchmark Models}
    We evaluate the performance of mainstream MLLMs on VIR-Bench, including both open-weight models (VideoLLaMA3~\cite{zhang2025videollama3frontiermultimodal}, LLaVA-Video~\cite{zhang2024videoinstructiontuningsynthetic}), InternVL3~\cite{zhu2025internvl3exploringadvancedtraining}, Qwen2.5-VL~\cite{bai2025qwen25vltechnicalreport}) and proprietary models (GPT-4.1~\cite{openai2025_gpt41_systemcard}, o4-mini~\cite{openai2025_o3_o4_mini_systemcard}, Gemini-2.5-Flash and Pro~\cite{comanici2025gemini25pushingfrontier}).
    All models are evaluated in a zero-shot setting.
    We use as many input frames as permitted by each model’s interface or pre-training setup; only the Gemini models accept audio input, and full details appear in Appendix~\ref{model-details}.
\subsection{Evaluation Metrics}
    We evaluate models using macro-averaged precision, recall, and F1 across both node and edge prediction.
    For prefecture and city nodes, a prediction is considered correct only if it exactly matches the gold label's surface name.
    For POIs, we apply a lightweight sequence-matching algorithm: predictions with a similarity score above 0.7 (high similarity) are treated as correct; predictions with a score above 0.5 (moderate similarity) are also accepted if the predicted POI category matches the gold category; all others are treated incorrect.
    For inclusion and transition edges, a prediction is counted as correct only when the tuple \texttt{<source, target, edge\_type>} exactly matches the gold tuple.
\subsection{Main Results}
    \begin{table}[t]
    \centering
    \small
    \resizebox{\linewidth}{!}{%
    \begin{tabular}{l|rrr|rrr|rrr|r}
    \toprule
    \textbf{Model} & \multicolumn{3}{c|}{\textbf{Prefecture}} & \multicolumn{3}{c|}{\textbf{City}} & \multicolumn{3}{c|}{\textbf{POI}} & \textbf{OVR} \\
     & Precision & Recall & F1 & Precision & Recall & F1 & Precision & Recall & F1 & F1 \\
    \midrule\\[-3.2ex]
    \multicolumn{11}{l}{\cellcolor[HTML]{EFEFEF}\textit{Open-weight}} \\
    VideoLLaMA3-7B & 29.8 & 41.6 & 31.6 & 19.1 & 14.4 & 14.7 & 24.0 & 10.3 & 13.7 & 14.6 \\
    LLaVA-Video-7B & 15.3 & 13.7 & 13.2 & 4.8& 6.5& 4.9& 10.8 & 4.7& 6.0& 5.9\\
    LLaVA-Video-72B & 17.8 & 24.7 & 18.5 & 10.6 & 10.4 & 9.1& 10.2 & 7.9& 7.8& 8.5\\
    InternVL3-8B & 20.1 & 17.3 & 17.9 & 8.8& 8.0& 7.2& 10.4 & 4.5& 6.0& 6.8\\
    InternVL3-38B & 48.6 & 46.3 & 45.7 & 29.8 & 19.6 & 21.8 & 19.4 & 11.5 & 13.6 & 16.4 \\
    InternVL3-78B & 58.2 & 60.5 & 56.7 & 41.1 & 33.7 & 33.5 & 30.4 & 14.8 & 19.0 & 22.8 \\
    Qwen2.5-VL-7B & 46.9 & 45.1 & 44.5 & 30.7 & 25.3 & 25.3 & 27.5 & 16.8 & 19.8 & 21.9 \\
    Qwen2.5-VL-32B & 74.7 & 70.6 & 69.7 & 53.6 & 38.1 & 41.2 & 37.4 & 26.1 & 29.2 & {\cellcolor[HTML]{CAFAC9}33.0} \\
    Qwen2.5-VL-72B & \textbf{86.2} & \textbf{73.6} & \textbf{77.2} & \textbf{65.4} & \textbf{43.8} & \textbf{49.0} & \textbf{52.3} & \textbf{26.6} & \textbf{33.9} & {\cellcolor[HTML]{34FF34}38.1} \\
    \\[-3.2ex]\midrule\\[-3.2ex]
    \multicolumn{11}{l}{\cellcolor[HTML]{EFEFEF}\textit{Proprietary}} \\
    GPT-4.1 & \textbf{91.2} & 85.9 & 86.5 & \textbf{75.9} & 62.6 & 66.0 & 61.0 & 51.0 & \textbf{53.6} & {\cellcolor[HTML]{CAFAC9}57.0} \\
    o4-mini & 90.3 & 85.6 & 86.1 & 71.3 & 59.0 & 62.3 & \textbf{63.1} & 44.8 & 50.4 & 53.9 \\
    Gemini-2.5-Flash & 88.7 & 85.5 & 85.1 & 74.3 & 63.7 & 65.6 & 57.0 & 50.4 & 51.5 & 55.3 \\
    Gemini-2.5-Pro & 89.7 & \textbf{89.0} & \textbf{87.7} & 73.4 & \textbf{68.2} & \textbf{68.6} & 51.8 & \textbf{58.1} & 52.8 & {\cellcolor[HTML]{34FF34}57.4} \\
    [-0.5ex]\bottomrule
    \end{tabular}}
    \vspace{5pt}
    \caption{Evaluation results on node prediction. ``OVR'' abbrev for ``Overall''. The open-weights and proprietary models with the highest and second-highest overall average scores are highlighted with bright green and light green marks.}
    \label{table:node-prediction}
    \end{table}
    \begin{figure*}[t]
    \captionsetup{type=table}
    \centering
    \small
        \begin{minipage}{0.68\linewidth} 
        \centering
        \resizebox{\linewidth}{!}{%
            \begin{tabular}{l|rrr|rrr|r}
            \toprule
            \textbf{Model} & \multicolumn{3}{c|}{\textbf{Inclusion}} & \multicolumn{3}{c|}{\textbf{Transition}} & \textbf{OVR} \\
             & Precision & Recall & F1 & Precision & Recall & F1 & F1 \\
            \midrule\\[-3.2ex]
            \multicolumn{8}{l}{\cellcolor[HTML]{EFEFEF}\textit{Open-weight}} \\
            VideoLLaMA3-7B & 39.8 & 31.1 & 33.4 & 2.5 & 1.2 & 1.4 & 23.9 \\
            LLaVA-Video-7B & 22.7 & 20.7 & 21.5 & 1.7 & 0.9 & 1.1 & 15.4 \\
            LLaVA-Video-72B & 66.3 & 60.3 & 62.5 & 10.3 & 8.6 & 8.4 & 42.4 \\
            InternVL3-8B & 48.3 & 45.2 & 46.0 & 4.9 & 2.4 & 2.5 & 31.2 \\
            InternVL3-38B & 64.7 & 59.7 & 61.9 & 15.6 & 12.3 & 12.9 & 41.8 \\
            InternVL3-78B & 74.4 & 67.5 & 70.6 & 20.7 & 11.8 & 13.4 & {\cellcolor[HTML]{CAFAC9}48.9} \\
            Qwen2.5-VL-7B & 32.0 & 28.6 & 29.6 & 1.5 & 1.3 & 1.3 & 18.0 \\
            Qwen2.5-VL-32B & 66.6 & 61.3 & 63.5 & 23.5 & 15.5 & 16.5 & 44.6 \\
            Qwen2.5-VL-72B &\textbf{76.5} & \textbf{69.2} & \textbf{72.3} & \textbf{32.6} & \textbf{18.5} & \textbf{20.8} & {\cellcolor[HTML]{34FF34}52.4} \\
            \\[-3.2ex]\midrule\\[-3.2ex]
            \multicolumn{8}{l}{\cellcolor[HTML]{EFEFEF}\textit{Proprietary}} \\
            GPT-4.1 & 78.3 & 75.9 & 76.5 & 34.2 & 36.2 & 34.4 & 58.8 \\
            o4-mini & 86.0 & 79.0 & 82.0 & 40.9 & 41.0 & 40.5 & {\cellcolor[HTML]{CAFAC9}64.9} \\
            Gemini-2.5-Flash & 83.1 & 74.9 & 78.5 & 42.8 & 42.4 & 41.7 & 63.4 \\
            Gemini-2.5-Pro & \textbf{94.8} & \textbf{87.6} & \textbf{90.8} & \textbf{66.4} & \textbf{68.0} & \textbf{66.8} & {\cellcolor[HTML]{34FF34}80.7} \\
            [-0.5ex]\bottomrule
            \end{tabular}}
            \caption{Evaluation results on edge prediction.}
            \label{table:edge-prediction}
        \end{minipage}
        \hfill
        \begin{minipage}{0.31\linewidth}
        \centering
        \vspace{12pt}
        \captionsetup{type=figure}
            \includegraphics[width=0.95\textwidth]{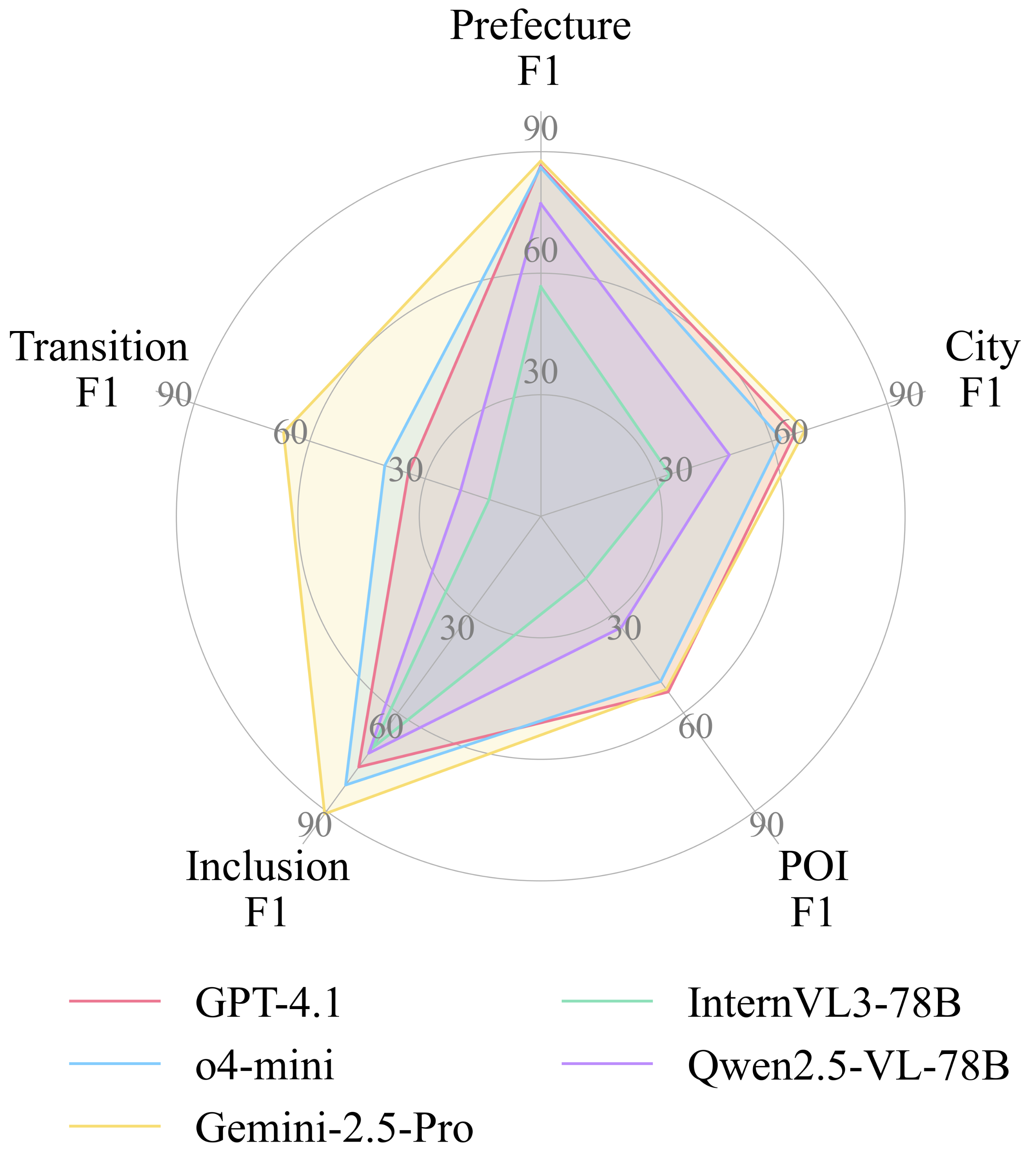}
            \vspace{12pt}
            \caption{Overall results of top-performing models.}
            \label{fig:edge-prediction}
        \end{minipage}
    \end{figure*}
    We present the evaluation results for node prediction in Table~\ref{table:node-prediction} and for edge prediction in Table~\ref{table:edge-prediction}.
    The overall scores are computed as weighted averages across different node and edge types, with weights proportional to the number of elements in each task category.
    We also provide a detailed error analysis in Appendix~\ref{error-analysis}.
    
    \paragraph{Overall Performance}
    Across all five task categories, open-weight models continue to underperform proprietary models.
    The strongest open model, Qwen2.5-VL-72B, comes close to proprietary performance on the easier categories (prefecture node prediction and inclusion edge prediction), but substantial gaps remain on the harder categories (POI node prediction and transition edge prediction).
    Other open models perform markedly worse: the LLaVA-Video series and InternVL3-8B achieve only single-digit F1 in city and POI node prediction, and five of the nine models also remain in single digits on transition edge prediction.
    In the proprietary models, Gemini-2.5-Pro is the top performer, especially on edge prediction, yet its F1 scores for city/POI node and transition edge prediction remain around 60.
    Taken together, these findings indicate that VIR-Bench is highly challenging for current MLLMs and highlight persistent limitations in geospatial and temporal understanding.
    
    \paragraph{Weak Results on Transition Edge Prediction}
    Across the tables, transition edge prediction is the most challenging task.
    Video-LLaMA3-7B, LLaVA-Video-7B, and Qwen2.5-VL-7B score only around 1, close to random guessing.
    A plausible factor is the limited number of input frames (e.g., 64 for the LLaVA-Video and InternVL3 series); however, models with larger budgets (180 for Video-LLaMA3 and 256 for Qwen2.5-VL) still struggle, suggesting the issue is not solely input length.
    Inspecting outputs from lower-performing models reveals two recurrent failure modes:
    (i) The model sometimes misinterprets the task, including the definitions of inclusion and transition edges; although it produces valid JSON, it yields nonsensical tuples such as \texttt{<Tokyo, Shibuya, transition>}.
    (ii) Transition edges are constrained to connect locations at the same hierarchical level; for POIs, edges are permitted only between POIs within the same city. Models often ignore this constraint and predict cross-city transitions, for example linking a POI in Tokyo to one in Osaka.
    These factors render the task an even more challenging test of temporal reasoning.
    
    \paragraph{Impact of Model Size}
    When comparing models of different sizes within the same families (LLaVA-Video, InternVL3, Qwen2.5-VL), we observe steady, scale-driven gains on node prediction, whereas edge prediction shows a sharp jump; for example, transition F1 improves by about $16\times$ from Qwen2.5-VL-7B to Qwen2.5-VL-72B.
    This pattern reflects the task demands: node prediction is a localized, single-point task that primarily relies on geospatial knowledge encoded in the models, whereas edge prediction requires a holistic view of the itinerary and thus benefits more from larger models with stronger long-context and temporal reasoning.
    An exception is LLaVA-Video-72B, which shows minimal improvement over LLaVA-Video-7B in POI node prediction.
    This is likely due to the limited geographic coverage in the LLaVA-OneVision training data~\cite{li2024llavaonevisioneasyvisualtask}.
    In contrast, models like Qwen2.5-VL demonstrate strong geo-localization capabilities; in our internal evaluations, Qwen2.5-VL was able to accurately predict POI coordinates, indicating extensive pretraining on geographic data.
    
    \paragraph{Thinking Models' Performance}
    Among the evaluated models, o4-mini and Gemini-2.5-Pro are the only ones that perform explicit ``thinking'' at inference.
    Although neither has an available non-thinking counterpart, we use GPT-4.1 and Gemini-2.5-Flash as proximate baselines to gauge what thinking contributes on VIR-Bench.
    On node prediction, the gains from thinking are limited: for POI nodes, o4-mini achieves higher precision but lower recall, while Gemini-2.5-Pro shows the opposite trend, suggesting different thinking strategies between OpenAI and Google.
    In contrast, for edge prediction, enabling deliberate thinking yields large gains for both models, especially Gemini-2.5-Pro, indicating that temporal understanding demands more complex reasoning.
    We presume that Gemini’s advantage stems from its use of audio, which supplies continuous, fine-grained temporal cues that sparsely sampled frames cannot provide, highlighting the need for truly multimodal modeling.
    To further validate the impact of reasoning and audio usage, we conduct additional ablation studies in the following section.
\subsection{Ablation Study}

\begin{table}[t]
\centering
\small
\setlength{\tabcolsep}{1mm}
\begin{tabular}{lcc|c|c}
\toprule
\textbf{Factor} & \textbf{Model}  & \textbf{Setting}& \textbf{Node (Prefecture/City/POI)} & \textbf{Edge (Inclusion/Transition)} \\
\midrule
\multirow{3}{*}{Frames} & \multirow{3}{*}{GPT-4.1} &64  & 85.8 / 62.5 / 39.6 & 76.6 / 27.6 \\
 && 128 & 85.4 / 64.0 / 52.9 & \textbf{78.8} / 33.5 \\
 && 256 & \textbf{86.5} / \textbf{66.0} / \textbf{53.6} & 76.5 / \textbf{34.4} \\
\midrule
\multirow{3}{*}{Reasoning}& \multirow{3}{*}{o4-mini}  & low    & \textbf{86.8} / 62.0 / 49.1 & 77.8 / 30.0 \\
 && medium & 86.1 / 62.3 / 50.4 & 82.0 / 40.5 \\
 && high   & 86.4 / \textbf{63.3} / \textbf{51.2} & \textbf{83.2} / \textbf{43.8} \\
\midrule
\multirow{2}{*}{Audio} & \multirow{2}{*}{\shortstack{Gemini-\\2.5-Flash}} & \ding{51} & \textbf{85.1} / \textbf{65.6} / \textbf{51.5} & 78.5 / \textbf{41.7} \\
 && \ding{55} & 82.5 / 64.0 / 50.5 & \textbf{82.6} / 22.3 \\
\bottomrule
\end{tabular}
\vspace{5pt}
\caption{Ablation results across frame count, reasoning effort, and audio usage. “Node” = Prefecture / City / POI nodes, and “Edge” = Inclusion / Transition edges. All reported scores are F1 values.}
\label{tab:ablation}
\end{table}

We conduct additional ablations varying the number of input frames, reasoning effort, and audio usage; results are reported in Table~\ref{tab:ablation}.
Increasing the number of input frames consistently improves GPT-4.1's overall performance.
In particular, the model limited to 64 frames performs poorly on POI node and transition edge prediction, suggesting that for videos in our benchmark, at least 128 frames ($\sim$1 frame every 14s) are a minimum requirement for reliable temporal reasoning.
Higher reasoning effort (i.e., longer thinking) leads o4-mini to better performance, especially on transition edge prediction, confirming our earlier observation that the task requires high-level, long-context reasoning.
Removing audio from Gemini-2.5-Flash yields worse results across most categories, with nearly a 50\% drop on transition edge prediction.
This confirms that audio is essential for temporal understanding, as it offers finer and more continuous granularity than the video stream (sampled at 1 fps), likely supporting more consistent long-context reasoning.
\section{Travel-planning Agent}
After watching a travel vlog, an animation, or a movie, many fans go on a pilgrimage: visiting the featured locations in the same order as they appear.
An automatically generated tranvel plan derived from the video and its visiting order graph would greatly streamline this process.

In this section, we construct MLLM-based agents which aim to provide travel plans given the videos.
The purposes of this experiment are (1) to demonstrate the importance of the itinerary reconstruction for this application and (2) to explore the feasibility of generating travel plans from videos, a capability not substantially addressed in prior work.

\subsection{Task Definition}
\paragraph{Input}
The agent system takes a list of POIs, a video and planning constraints as input.
While this list of POIs could be the output of the node prediction task, we use the POI list from the video annotation in this experiment to isolate the evaluation from model performance.
The video provides richer information for the planning process.
The constraints consist of the number of companions, travel duration and travel budget inspired by previous work \cite{xie2024travelplanner}.

\paragraph{Output}
The output is a travel plan formatted in Markdown.
It includes basic information such as the destination, duration, and budget.
A core component is a detailed day-by-day itinerary, specifying a schedule of activities, visiting times for each POI, and transportation methods.
This itinerary is supplemented with POI details and other rich information extracted from the provided video and/or the search results.
Furthermore, the plan provides practical recommendations, including specific restaurants and accommodations with relevant details like price and ratings.

\subsection{Implementation of the Agent System}
We implement the system as a multi-agent framework coordinated by an autonomous orchestrator. 
This central component is responsible for dynamically determining the execution order of agents, managing the shared state of them.

The framework comprises five specialized agents, each tasked with a specific function:
\textbf{Plan Agent} constructs the day-by-day schedule, optimizing time allocation based on the user's budget and constraints.
\textbf{Google Maps Agent} retrieves POI details.
\textbf{Route Agent} finds the routes between POIs given the list of POIs.
\textbf{Accommodation Agent} finds suitable lodging that fits the budget and is optimally located relative to the planned activities.
\textbf{Summary Agent} integrates the outputs from all other agents to generate a unified final report, including a complete travel plan and a budget breakdown.
Each agent can use tools that are appropriate for its purpose, including Google Maps API-based tools and browser-based tools.
Further implementation details are shown in Appendix~\ref{agent-implementation-details}.

\subsection{Experimental Setup}
\subsubsection{System Setup}
To verify the importance of the itinerary reconstruction step, the core task of our benchmark, we prepare 3 input settings: a list of POIs only (\textbf{POI}), a video only (\textbf{Video}) and both of them (\textbf{P+V}).
Constraints are always provided as input in all settings.

The backbone models of the orchestrator and all agents are fixed as Gemini-2.5-Pro.
For the reproducibility, the temperature is set as zero.

\subsubsection{Evaluation Setup}
We sample 20 pairs of videos and their corresponding annotated graphs as input.
Agents under the three settings then generate travel plans for each pair, resulting in 60 plans for evaluation.

We qualitatively evaluate the generated plans with crowdsourcing\footnote{\url{https://crowdsourcing.yahoo.co.jp}}.
Since the videos are filmed across Japan, we hire Japanese-speaking crowdworkers and translate the plans into Japanese.
Each plan is evaluated by five workers.
The evaluation consists of four tasks: assessing the plan's attractiveness (\textbf{Attraction}), verifying the feasibility of the transportation information (\textbf{Feasibility}), judging the suitability of the number of POIs (\textbf{Density}), and determining the plan's consistency with the video (\textbf{Alignment}).

We also evaluate the system's POIs selection.
Using GPT-4o~\cite{openai2024gpt4ocard}, we extract the POIs mentioned in the generated plans and treat them as selected.
We then compare selected versus unselected POIs in terms of their on-screen duration in the video and their Google Maps ratings.

\begin{figure*}[t]
  \centering
  \subfloat[Attraction Task]{
    \includegraphics[width=0.35\textwidth]{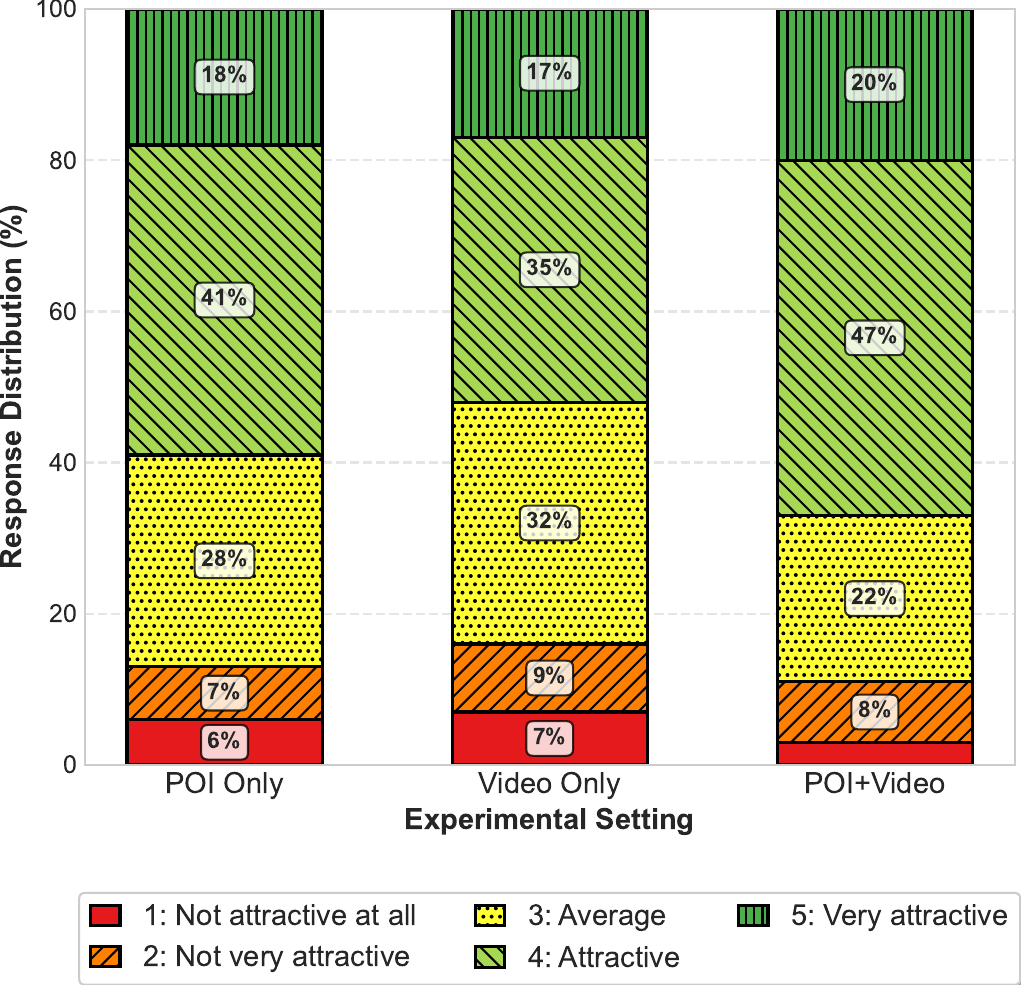}
    \label{fig:cs_attract}
  }
  \subfloat[Feasibility Task]{
    \includegraphics[width=0.35\textwidth]{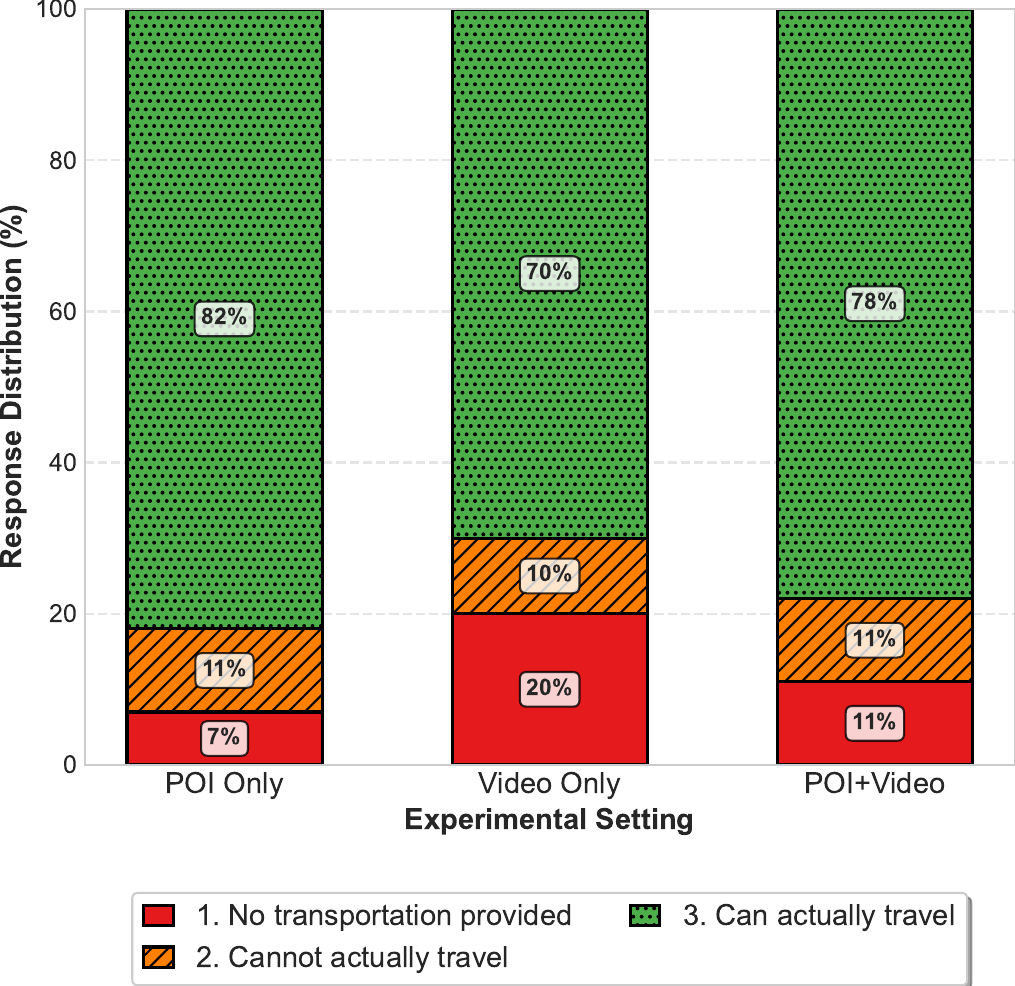}
    \label{fig:cs_transport}
  }
  \\
  \subfloat[Density Task]{
    \includegraphics[width=0.35\textwidth]{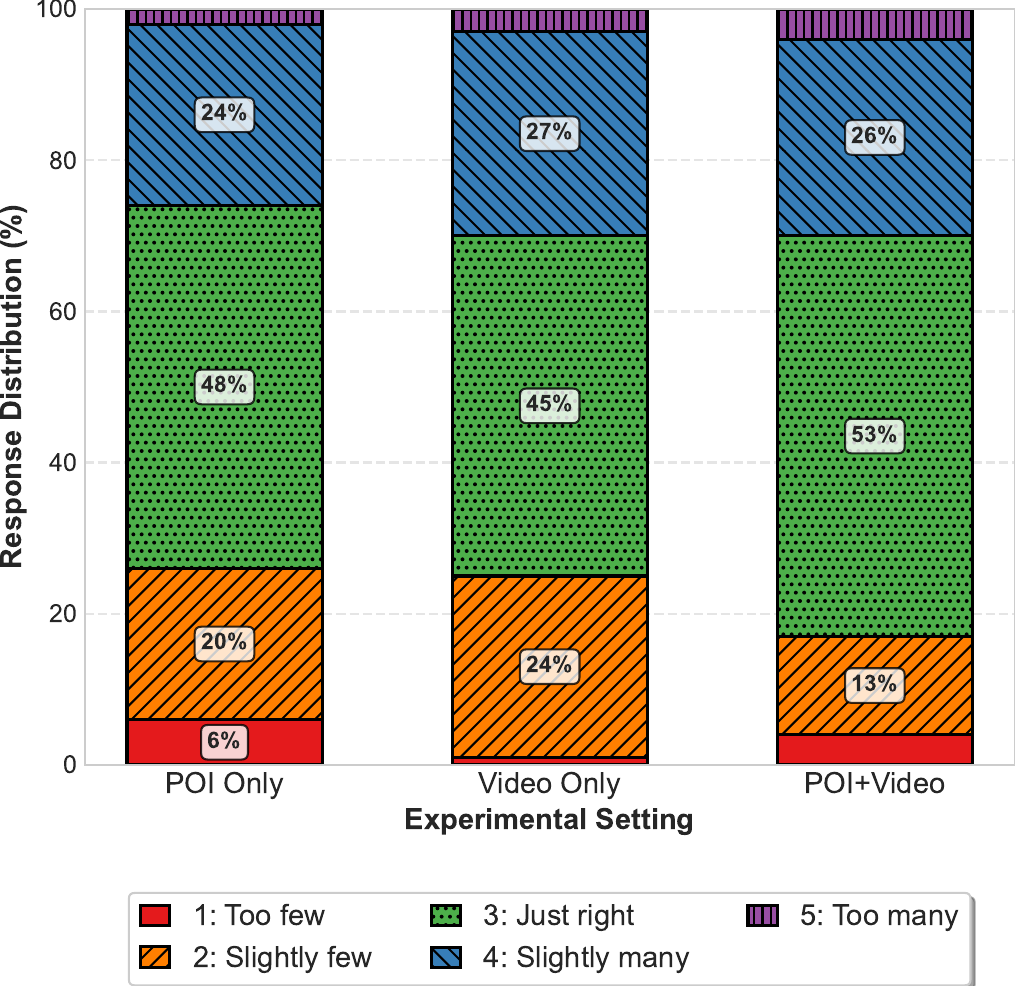}
    \label{fig:cs_density}
  }
  \subfloat[Alignment Task]{
    \includegraphics[width=0.35\textwidth]{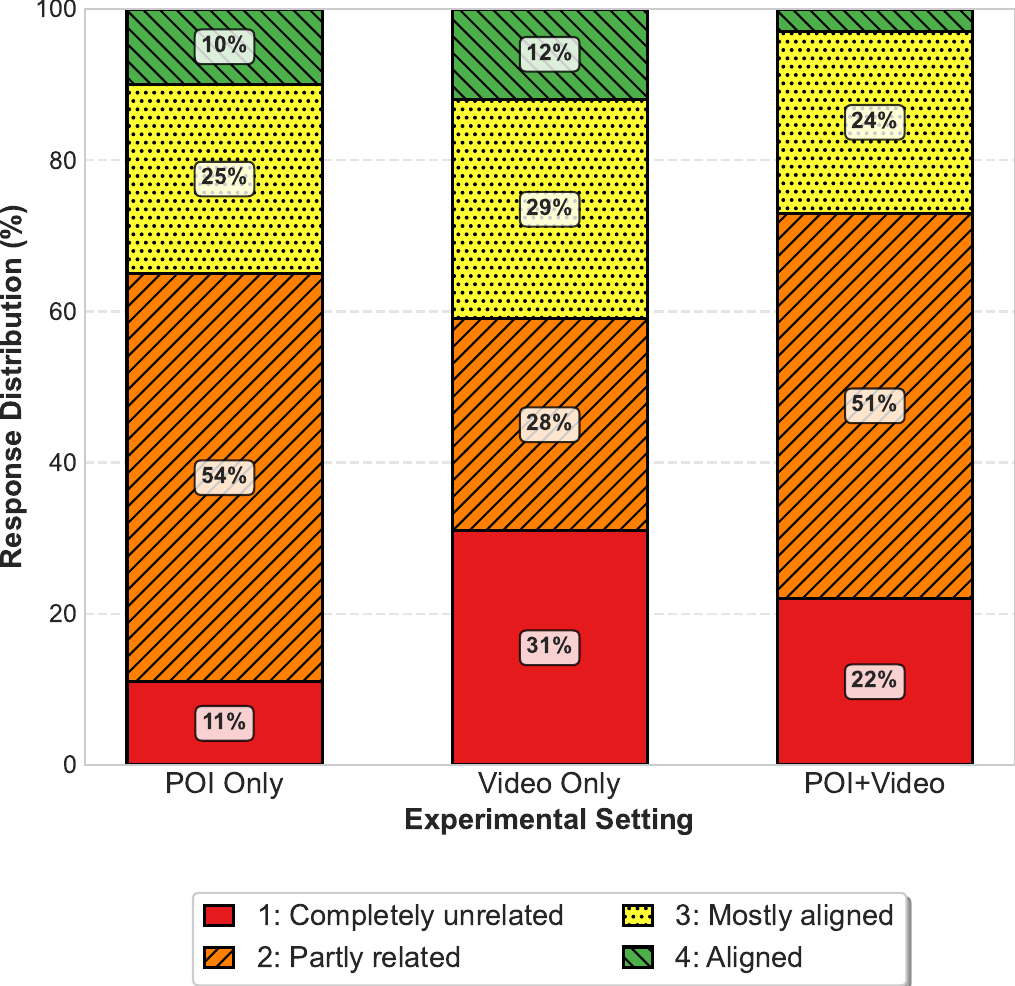}
    \label{fig:cs_relevance}
  }
  \caption{Crowdsourcing results of the agent system.}
  \label{fig:agent_cs}
\end{figure*}

\subsection{Results and Discussion}

\paragraph{Crowdsourcing Evaluation}
The crowdsourcing results are summarized in Figure~\ref{fig:agent_cs} and Table~\ref{tab:crowdsourcing_results}.
These results indicate that the P+V setting yielded the most attractive travel plans, achieving the highest average score of 3.73 in the Attraction task (Table~\ref{tab:crowdsourcing_results}).
This suggests that while a list of POIs provides a solid foundation, the rich information from the video—such as the atmosphere of a place or specific activities shown—is crucial for creating a more appealing plan.

The Alignment task reveals the critical challenge of POI extraction from video (Figure~\ref{fig:cs_relevance}).
The video-only setting produced the most polarized results: while it achieved the highest proportion of ``mostly aligned'' or ``aligned'' plans (41\%), it also generated the largest share of plans deemed ``completely unrelated'' (31\%).
This instability suggests that the agent's success is highly dependent on the initial, error-prone step of identifying POIs from raw video.
A failure in this stage, as evidenced by the low node prediction F1-scores (Table~\ref{table:node-prediction}), leads directly to a final plan that is misaligned with the video's content.
This underscores that a robust itinerary reconstruction is a foundational prerequisite for generating contextually relevant and reliable travel plans.

In terms of practicality, all settings generated feasible plans.
The POI-only setting was most reliable in providing transportation information, while the video-based settings were slightly better at creating a plan with a ``just right'' density of activities (Figure~\ref{fig:cs_density}).
The video-only setting had the highest proportion of plans lacking transportation information (22\%, Figure~\ref{fig:cs_transport}) among the three settings.

\paragraph{Analysis of POI Selection}
Table~\ref{tab:poi_detailed_statistics} provides deeper insights into the agent's underlying POI selection strategy.
For all input settings, the agent showed a strong and statistically significant tendency to select POIs that were featured for a longer duration in the video.
This effect was most pronounced in the P+V setting ($\Delta=+41.7,p<0.001$), suggesting that the combination of a POI list and video context enables the agent to most effectively identify and prioritize key locations.

Furthermore, the agent also showed a preference for POIs with higher Google Maps ratings, although this effect was less pronounced than that of video duration.
These findings indicate that the agent intelligently synthesizes signals from both the video (visual prominence) and external knowledge sources (user ratings) to make its planning decisions.

\begin{table}[t]
\centering
\small
\setlength{\tabcolsep}{1mm}
\begin{tabular}{l|ccccc}
\toprule
\multirow{3}{*}{\textbf{Input}} & \multicolumn{3}{c}{\textbf{Mean Score}} & \multicolumn{2}{c}{\textbf{Transportation} (\%)} \\
\cmidrule(lr){2-4} \cmidrule(lr){5-6}
 & Attract & Density & Relevance & \multirow{2}{*}{Has Info} & \multirow{2}{*}{Feasible} \\
 & (1-5) & (1-5)$^\dagger$ & (1-4) & & \\
\midrule
POI      & 3.58 & \textbf{2.96} & \textbf{2.34} & \textbf{93.0} & \textbf{88.2} \\
Video    & 3.46 & 3.07 & 2.22 & 80.0 & 87.5 \\
P+V & \textbf{3.73} & 3.13 & 2.08 & 89.0 & 87.6 \\
\bottomrule
\multicolumn{6}{l}{\footnotesize $^\dagger$ For Density: 1=too little, 3=just right, 5=too much.} \\
\end{tabular}
\caption{Crowdsourcing results by system configuration.}
\label{tab:crowdsourcing_results}
\end{table}



\begin{table}[t]
    \centering
    \small
    \setlength{\tabcolsep}{0.5mm}
    \begin{tabular}{l|cccccc}
        \toprule
        \multirow{2}{*}{\textbf{Input}}& \multicolumn{3}{c}{\textbf{Duration} (seconds)} & \multicolumn{3}{c}{\textbf{Google Rating} (1-5)} \\
        \cmidrule(lr){2-4} \cmidrule(lr){5-7}
         & Selected & Unselected & $\Delta$ & Selected & Unselected & $\Delta$ \\
        \midrule
        POI          & 58.4& 36.8& 21.6$^{\dagger \dagger \dagger}$ & 4.29& 4.17& \textbf{0.12}$^{\dagger \dagger \dagger}$ \\
        Video        & 68.6& 34.4& 34.2$^{\dagger \dagger \dagger}$ & 4.26& 4.19& 0.07$^\dagger$ \\
        P+V          & 76.3& 34.7& \textbf{41.7}$^{\dagger \dagger \dagger}$ & 4.25& 4.19& 0.06$^\dagger$ \\
        \midrule
        Total        & 57.2& 32.6& 24.6$^{\dagger \dagger \dagger}$ & 4.25& 4.17& 0.08$^{\dagger \dagger}$ \\
        \bottomrule
        \multicolumn{7}{l}{\footnotesize $\dagger: p < 0.05$, $\dagger \dagger: p < 0.01$, $\dagger \dagger \dagger: p < 0.001$}
    \end{tabular}
    \caption{
        Comparison of statistics for selected vs. unselected POIs by system configuration.
        $\Delta$ denotes the difference (``selected'' minus ``unselected''). 
        In the ``Total'' row, ``selected'' refers to POIs chosen in at least one setting, and ``unselected'' refers to POIs never chosen in any setting.
    }
    \label{tab:poi_detailed_statistics}
\end{table}

\section{Conclusion}
We presented VIR-Bench, a video understanding benchmark designed to evaluate long-range geospatial–temporal reasoning through itinerary reconstruction, utilizing visiting order graphs constructed from 200 travel videos.
By decomposing the task into node and edge prediction, we revealed persistent weaknesses in state-of-the-art MLLMs: open-weight models consistently lag behind proprietary ones, and transition edges remain the primary bottleneck. 
Our prototype travel-planning agent further illustrated the practical value of VIR-Bench, demonstrating that combining POI lists with videos generates the most appealing travel plans, particularly emphasizing visually salient and highly rated POIs.
In summary, VIR-Bench provides both a rigorous benchmark and a practical foundation for advancing MLLMs toward video-grounded geospatial–temporal understanding in real-world travel planning scenarios.
Looking ahead, we intend to expand our dataset by increasing geographic and linguistic diversity, incorporating a broader range of filming styles, and enhancing annotation richness.
Furthermore, we aim to explore advanced agent systems capable of referencing multiple videos simultaneously, enabling the generation of more comprehensive and engaging travel plans.

\bibliographystyle{custom}
\bibliography{ref}

\appendix
\etocsettocstyle{\section*{Appendix Contents}}{}
\begin{CJK}{UTF8}{ipxm}
\input{appendix}
\end{CJK}

\end{document}

%% file: appendix.tex
\clearpage
\section{Benchmark Details}
\label{benchmark-details}
\subsection{Annotation Guidelines}
This section presents the complete annotation guidelines used in constructing VIR-Bench.

\subsubsection{Introduction}
This project constructs a video understanding benchmark to evaluate a multimodal LLM's temporal and spatial grounding capabilities.
Concretely, the task is to input a travel vlog into the MLLM and have it infer the subject's movement trajectory (e.g., \emph{Haneda Airport $\rightarrow$ Shinjuku Station $\rightarrow$ McDonald's Seibu-Shinjuku Ekimae}).
To enable trajectory analysis and evaluation, this annotation collects the set of POIs visited in each video.

The overall workflow consists of two stages: (1) selecting videos, and (2) annotating trajectories.

\subsubsection{Video Selection}
Please search YouTube for videos that satisfy all of the following required criteria:
\begin{enumerate}\small
  \item The video is a travel vlog set \textbf{within Japan}. Videos that include leaving Japan (e.g., a trip to Korea) are not allowed.
  \item The spoken language in the video is the language specified by the authors (Japanese or English).
  \item Apart from the intro/outro, the video progresses \textbf{in chronological order}. Narratives that jump back to earlier scenes are not allowed.
  \item The subjects do not split into multiple groups acting in parallel (no scenes where different places are visited simultaneously).
  \item It is \textbf{tourism-oriented}, visiting multiple POIs, rather than a simple neighborhood stroll.
  \item The video does \textbf{not} continuously overlay POI names as on-screen captions (short captions for a few seconds are acceptable).
  \item The preferred duration is \textbf{10--20 minutes}, and at most \textbf{30 minutes}.
\end{enumerate}
In addition to the required criteria, please aim to satisfy the following desirable criteria where possible:
\begin{enumerate}\small
  \item Geographic diversity across the set of videos assigned to a single annotator (e.g., avoid selecting ten videos all in Tokyo).
  \item Prefer videos that span multiple prefectures.
\end{enumerate}

\noindent
Example videos (for reference):
\begin{itemize}\small
  \item \url{https://www.youtube.com/watch?v=f-t3IFu-U7U}
  \item \url{https://www.youtube.com/watch?v=VjWr1NYVGPc}
  \item \url{https://www.youtube.com/watch?v=CuEOnd5cAA8}
  \item \url{https://www.youtube.com/watch?v=xqbiBTgcaiA}
\end{itemize}

\subsubsection{Trajectory Annotation}
\paragraph{Definition of a POI}
In this project, a POI is defined as a place registered on Google Maps. Exclude locations with extremely few reviews (0 or near 0), and places unrelated to tourism (e.g., a station-front public restroom).

\paragraph{Annotation Tool}
We use Vidat~\cite{zhang2020vidat}, a browser-based video annotation tool. Before starting, we strongly recommend watching \href{https://www.youtube.com/watch?v=BpjKy7FKajI}{the official 7-minute tutorial video}.

\paragraph{Procedure}
Annotate the video trajectory as follows:
\begin{enumerate}\small
  \item Access Vidat in your browser: \url{https://vidat2.davidz.cn/\#/}
  \item Load the downloaded video.
  \item For each POI visit, use the timeline to select the start and end time and create a new segment.
  \item Copy the corresponding Google Maps URL for that POI into the segment's \texttt{description}.\\
        Example: \url{https://maps.app.goo.gl/DT4ENgf8dUopWNUK6}.
  \item Move on to the next POI and repeat steps 3--4.
  \item When finished, save the annotation as \texttt{<videoID>.json}.
\end{enumerate}

\subsubsection{FAQ}
\small
\textbf{1. POIs with the Same Name}\\
\textbf{Q:} There are multiple entries named ``Sakuragichō Station.'' Which one?\\
\textbf{A:} Choose the entry with the larger number of reviews.
\\

\noindent
\textbf{2. Hierarchical Structure of POIs}\\
\textbf{Q:} What if a POI contains another POI? (e.g., Ghibli Park is within Expo 2005 Aichi Commemorative Park.)\\
\textbf{A:} Annotate all levels of the hierarchy. The parent POI's start/end times must enclose the child POIs' times. If the parent times are $(x_1,y_1)$ and the child times are $(x_2,y_2)$, ensure $x_1 < x_2 < y_2 < y_1 \ $.
\\

\noindent
\textbf{2.1 Minimum granularity of a POI.}\\
\textbf{Q:} Should I annotate a gift shop inside an aquarium?\\
\textbf{A:} A gift shop is a \emph{part facility} of the aquarium and need not be annotated. In principle, treat a building as the minimum POI unit. \emph{Exception:} in large \emph{multi-tenant malls} (e.g., AEON Mall), individual venues such as cinemas or Starbucks should be annotated.
\\

\noindent
\textbf{2.2 Maximum granularity of a POI.}\\
\textbf{Q:} Does ``Okinawa'' count as a POI?\\
\textbf{A:} ``Okinawa,'' ``Kokusai-dōri,'' and ``Minatomirai'' are areas on Google Maps (no review counts) and should not be annotated. However, ``Kokusai-dōri Shopping Street'' and ``Yokohama Chinatown'' are POIs (with reviews) and should be annotated.
\\

\noindent
\textbf{3. Concurrent Visits}\\
\textbf{Q:} What if multiple POIs are visited simultaneously?\\
\textbf{A:} Create multiple segments with the same start/end times.
\\

\noindent
\textbf{4. Unknown Venue Names}\\
\textbf{Q:} They visit a cafe, but I cannot identify which one.\\
\textbf{A:} Leave the URL blank and write \texttt{UNKNOWN} + the POI category. If you are unsure of the category, please consult. This is a last resort; keep the number of \texttt{UNKNOWN} entries as low as possible.
\begin{itemize}
  \item POI category list: \href{https://developers.google.com/maps/documentation/places/web-service/place-types?_gl=1*10we1b2*_up*MQ..*_ga*NzYzMDgxOTE2LjE3NDM4MzMyMDE.*_ga_NRWSTWS78N*MTc0MzgzMzIwMS4xLjEuMTc0MzgzMzIyMy4wLjAuMA..#table-a}{Google Maps Platform Place Type (Table A).}
  \item Examples: \texttt{UNKNOWN cafe}, \texttt{UNKNOWN ramen\_restaurant}, \texttt{UNKNOWN hotel}.
\end{itemize}

\noindent
\textbf{5. Chain Stores with Unknown Branch}\\
\textbf{Q:} It's a Starbucks, but I can't tell which branch.\\
\textbf{A:} Pick one plausible branch and paste its URL, then append \texttt{, branch unknown}.
\begin{itemize}
    \item Example: \url{https://maps.app.goo.gl/uSybfW2HRjJwrfrg8}\texttt{, branch unknown}
\end{itemize}

\noindent
\textbf{6. POIs Missing from Google Maps}\\
\textbf{Q:} A rural shop may not exist as a POI on Google Maps.\\
\textbf{A:} Leave the URL blank and write \texttt{<shop name> + <category>}, analogous to the \texttt{UNKNOWN} handling.
\\

\noindent
\textbf{7. Closed or Relocated Venues}\\
\textbf{Q:} The venue visited in the video has since closed or moved.\\
\textbf{A:} If it is \emph{closed} but still exists as a POI entry on Google Maps, annotate as usual. For \emph{relocations}, annotate according to the current location as listed on Google Maps.
\\

\noindent
\textbf{8. Intros/Outros}\\
\textbf{Q:} The intro (or outro) shows all POIs. Annotate them?\\
\textbf{A:} Annotate only the main content of the video.
\\

\noindent
\textbf{9. Start/End Time Definitions}\\
\textbf{Q:} How do we define the start and end times?\\
\textbf{A:} The start time is when the subject physically arrives at the POI; the end time is when they leave.\\[3pt]
\textbf{Q:} How precise must we be at hard cuts between scenes?\\
\textbf{A:} Do not annotate with frame-level strictness; reasonable precision is sufficient.
\\

\noindent
\textbf{10. Definition of ``Visit''}\\
\textbf{Q:} Do they need to enter the facility to count as a visit?\\
\textbf{A:} If they reach the front of the facility and it is visibly captured in the frame, count it as a visit. Distant landmarks merely visible on the horizon do not count.
\normalsize
\subsection{Statistics}
\begin{figure*}[htbp]
    \centering
    \subfloat[Distribution of unique prefectures per video\label{fig:unique_areas_dist}]{%
        \includegraphics[width=0.32\textwidth]{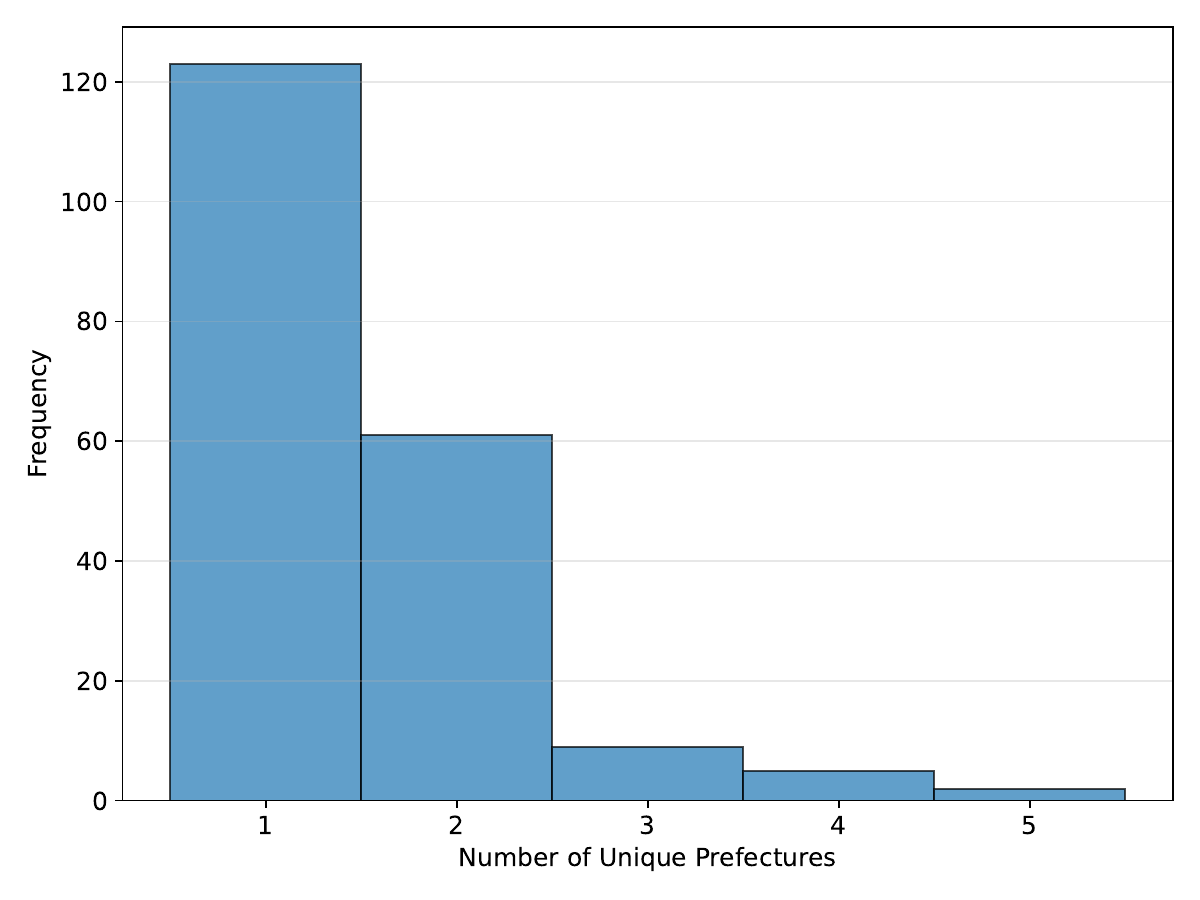}
    }
    \hfill
    \subfloat[Distribution of POI count per video\label{fig:poi_count_dist}]{%
        \includegraphics[width=0.32\textwidth]{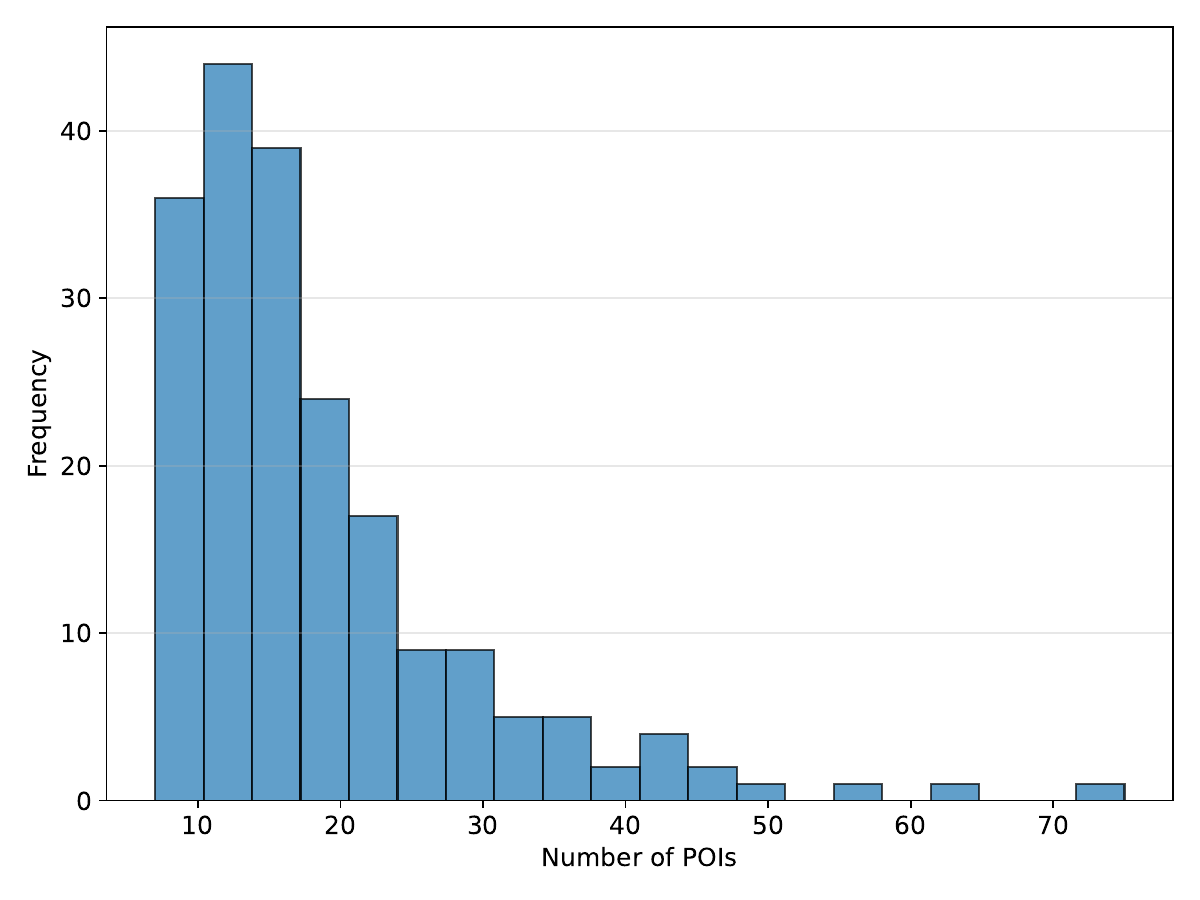}
    }
    \hfill
    \subfloat[Distribution of POI duration\label{fig:poi_duration_dist}]{%
        \includegraphics[width=0.32\textwidth]{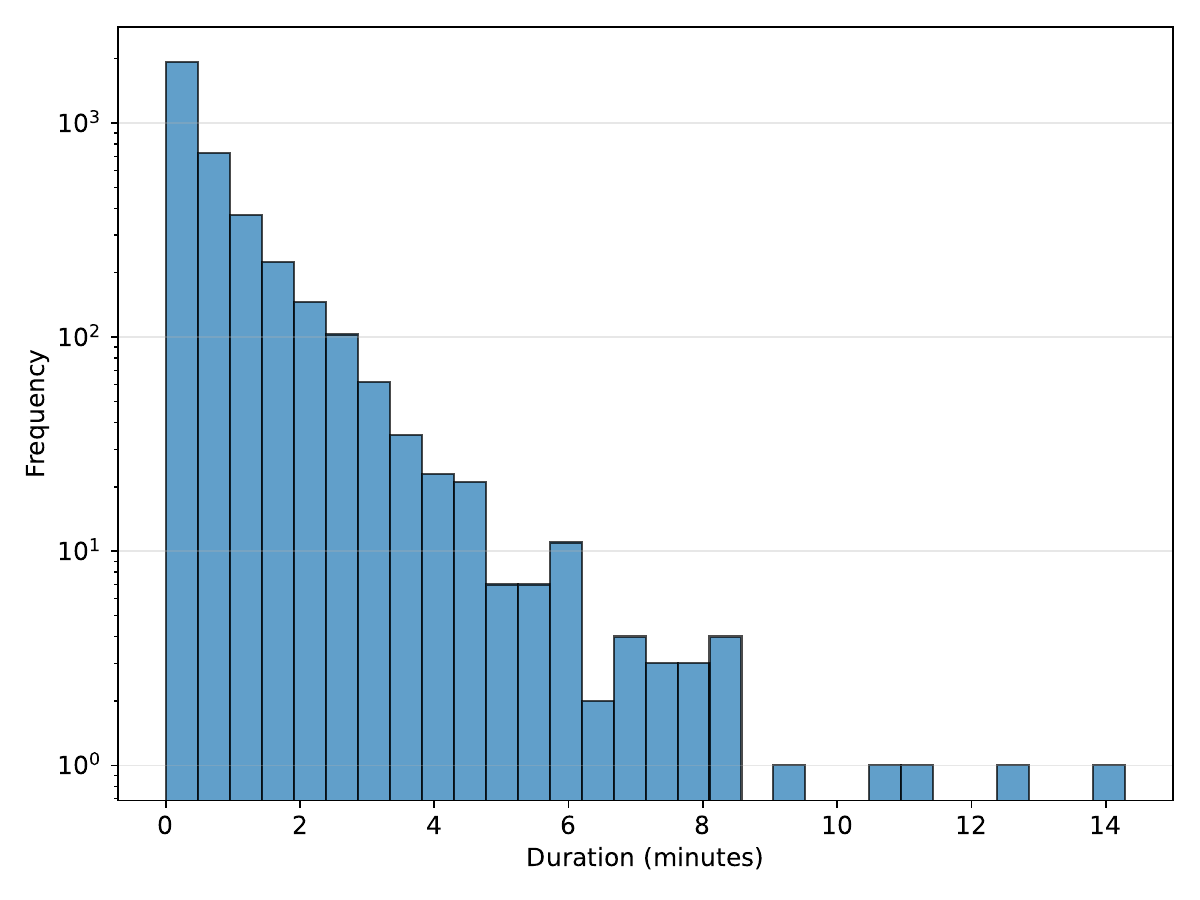}
    }
    \caption{Statistical distributions of the dataset. (a) shows the distribution of unique prefectures covered per video, with most videos spanning 1-2 prefectures. (b) presents the distribution of POI count per video, ranging from 7 to 75 POIs with a mean of 18.4. (c) displays the distribution of POI durations in minutes, showing a right-skewed distribution with most POIs lasting under 2 minutes.}
    \label{fig:dataset_statistics}
\end{figure*}

\begin{figure}[ht]
    \centering
    \includegraphics[width=0.7\linewidth]{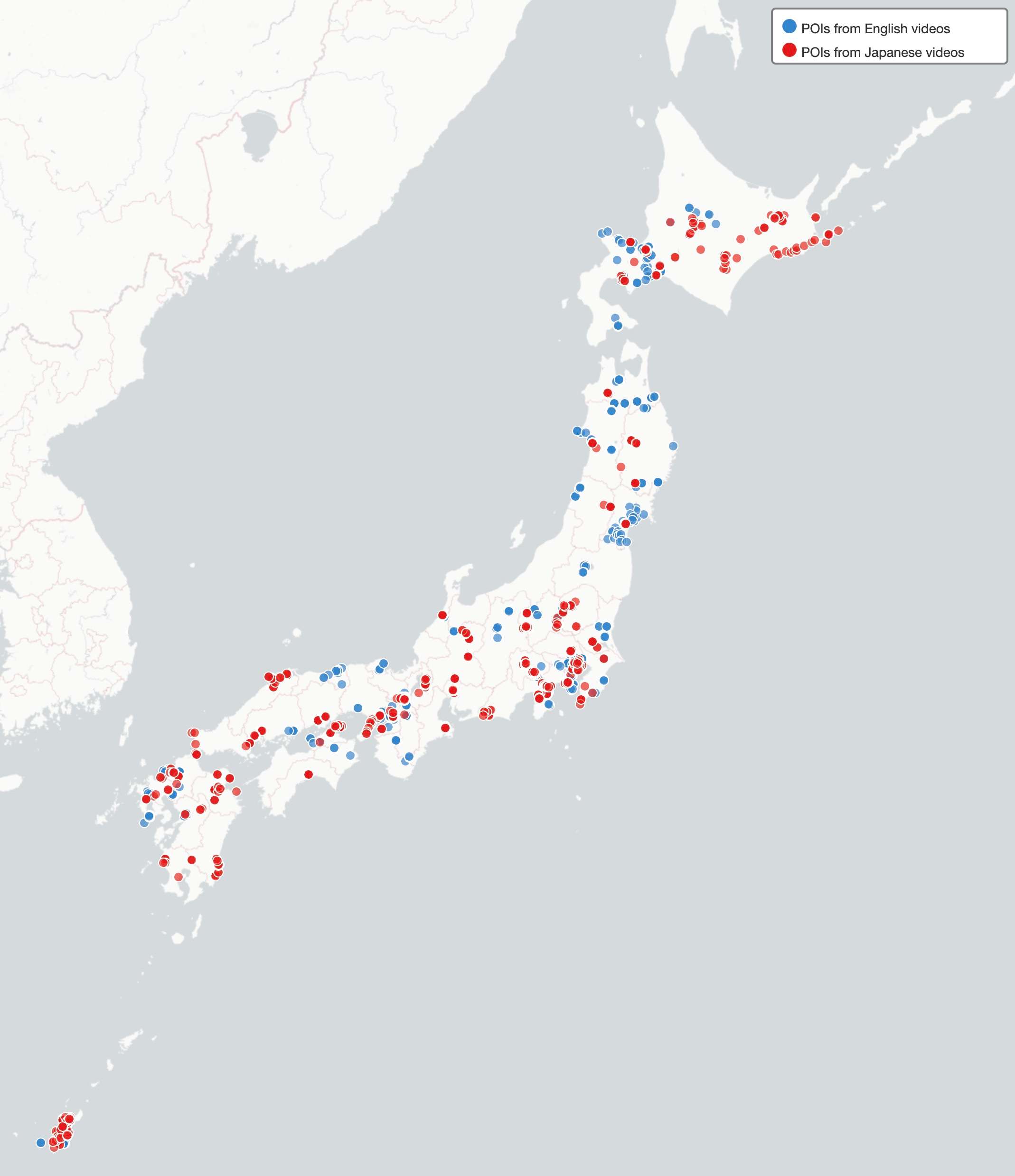}
    \caption{Geographical distribution of POIs across Japan, showing an even spread from Hokkaido in the northeast to Okinawa in the southwest. Blue points indicate POIs from English-speaking videos, while red points represent POIs from Japanese-speaking videos.}
    \label{fig:poi-map}
\end{figure}

VIR-Bench comprises 200 travel videos filmed across Japan, each paired with a corresponding visiting order graph that captures the itinerary depicted in the video.
Of these, 100 videos are narrated in English, while the rest are in Japanese.
Detailed statistics are shown in Table~\ref{tab:summary_stats}.
Figure~\ref{fig:dataset_statistics} presents key dataset statistics, while Figure~\ref{fig:poi-map} visualizes the geographical distribution of POIs across Japan.

\begin{table*}[ht]
\centering
\resizebox{\linewidth}{!}{
\begin{tabular}{l|cccc|cccc|cccc|ccc}
\toprule
\multirow{2}{*}{\textbf{Language}}& \multicolumn{4}{c|}{\textbf{\# POIs / video}} & \multicolumn{4}{c|}{\textbf{Video Duration (s)}} & \multicolumn{4}{c|}{\textbf{POI Duration (s)}} & \multicolumn{3}{c}{\textbf{\# Unique Prefectures}} \\
 & Min & Mean $\pm$ SD & Max & Sum & Min & Mean $\pm$ SD & Max & Sum & Min & Mean $\pm$ SD & Max & Sum & Min & Mean $\pm$ SD & Max \\
\midrule
Japanese & 7 & $15.4$ \scriptsize{$\pm 6.4$} & 36 & 1,544& 624.4 & $1061.2$ \scriptsize{$\pm 293.6$} & 1900.9 & 106,115 & 1.0 & $62.1$ \scriptsize{$\pm 78.1$} & 857.3 & 95,849 & 1& $1.4$ \scriptsize{$\pm 0.7$} & 4\\
English & 7 & $21.4$ \scriptsize{$\pm 13.1$} & 75 & 2,145& 433.4 & $1059.5$ \scriptsize{$\pm 299.3$} & 1707.6 & 105,945 & 0.4 & $42.4$ \scriptsize{$\pm 55.2$} & 658.2 & 90,702 & 1& $1.7$ \scriptsize{$\pm 0.9$} & 5\\
All & 7 & $18.4$ \scriptsize{$\pm 10.7$} & 75 & 3,689& 433.4 & $1060.3$ \scriptsize{$\pm 295.7$} & 1900.9 & 212,061 & 0.4 & $50.7$ \scriptsize{$\pm 66.5$} & 857.3 & 186,551 & 1& $1.5$ \scriptsize{$\pm 0.8$} & 5\\
\bottomrule
\end{tabular}}
\caption{Statistics of VIR-Bench.}
\label{tab:summary_stats}
\end{table*}

\subsection{License and Access}
We release the VIR-Bench dataset strictly for research purposes, in compliance with Article 30-4 (Use for Non-Enjoyment Purposes) and Article 47-5 (Minor Use in Information Analysis Services) of the Japanese Copyright Act. Commercial use of any kind is strictly prohibited. The dataset may not be redistributed on servers outside Japan or under alternative licenses.
\clearpage
\section{Experiment Details}
\subsection{Model Cards}
\label{model-details}
We list the models and their corresponding parameters used in this paper in Table~\ref{tab:model-cards}.

\begin{table*}[ht]
\centering
\small
\resizebox{\linewidth}{!}{%
\begin{tabular}{lrrcll}
\hline
\textbf{Model} & \textbf{Max Frames} & \textbf{Thinking Budget} & \textbf{Audio Input} & \textbf{Reference} & \textbf{Identifier / Repository} \\
\hline
VideoLLaMA3-7B & 180 & -& - & \cite{zhang2025videollama3frontiermultimodal} & \href{https://huggingface.co/DAMO-NLP-SG/VideoLLaMA3-7B}{DAMO-NLP-SG/VideoLLaMA3-7B} \\
LLaVA-Video-7B & 64 & - & -& \cite{zhang2024videoinstructiontuningsynthetic} & \href{https://huggingface.co/lmms-lab/LLaVA-Video-7B-Qwen2}{lmms-lab/LLaVA-Video-7B-Qwen2} \\
LLaVA-Video-72B & 64 & - & -& same as above & \href{https://huggingface.co/lmms-lab/LLaVA-Video-72B-Qwen2}{lmms-lab/LLaVA-Video-72B-Qwen2} \\
InternVL3-8B & 64 & - & -& \cite{zhu2025internvl3exploringadvancedtraining} & \href{https://huggingface.co/OpenGVLab/InternVL3-8B}{OpenGVLab/InternVL3-8B} \\
InternVL3-38B & 64 & - & -& same as above & \href{https://huggingface.co/OpenGVLab/InternVL3-38B}{OpenGVLab/InternVL3-38B} \\
InternVL3-78B & 64 & - & -& same as above & \href{https://huggingface.co/OpenGVLab/InternVL3-78B}{OpenGVLab/InternVL3-78B} \\
Qwen2.5-VL-7B & 256 & - & -& \cite{bai2025qwen25vltechnicalreport} & \href{https://huggingface.co/Qwen/Qwen2.5-VL-7B-Instruct}{Qwen/Qwen2.5-VL-7B-Instruct} \\
Qwen2.5-VL-32B & 256 & - & -& same as above & \href{https://huggingface.co/Qwen/Qwen2.5-VL-32B-Instruct}{Qwen/Qwen2.5-VL-32B-Instruct} \\
Qwen2.5-VL-72B & 256 & - & -& same as above & \href{https://huggingface.co/Qwen/Qwen2.5-VL-72B-Instruct}{Qwen/Qwen2.5-VL-72B-Instruct} \\
\hline
GPT-4.1 & 256 & - & - & \cite{openai2025_gpt41_systemcard} & \texttt{gpt-4.1-2025-04-14} \\
o4-mini & 256 & 32768 & -& \cite{openai2025_o3_o4_mini_systemcard} & \texttt{o4-mini-2025-04-16} \\
Gemini-2.5-Flash & 1 fps & - & 1 kbps & \cite{comanici2025gemini_full} & \texttt{gemini-2.5-flash} \\
Gemini-2.5-Pro & 1 fps & 32768 & 1 kbps & same as above & \texttt{gemini-2.5-pro} \\
\hline
\end{tabular}%
}
\caption{Models used in this paper.}
\label{tab:model-cards}
\end{table*}
\subsection{Error Analysis}
\label{error-analysis}
We categorize model errors into three types: (1) \textbf{Prompt Analysis Error}, where the model misinterprets the task or fails to follow instructions; (2) \textbf{Geographic Knowledge Error}, reflecting limited geographic knowledge (i.e., the ability to play GeoGuessr) encoded in the model; (3) \textbf{Temporal Reasoning Error}, stemming from flawed geospatial-temporal reasoning based on the video input.
Prompt analysis errors can arise in both node and edge prediction tasks, while geographic knowledge errors are primarily observed in node prediction, and temporal reasoning errors occur exclusively in edge prediction.
\subsubsection{Prompt Analysis Error}
\begin{figure}[ht]
    \centering
    \includegraphics[width=0.97\linewidth]{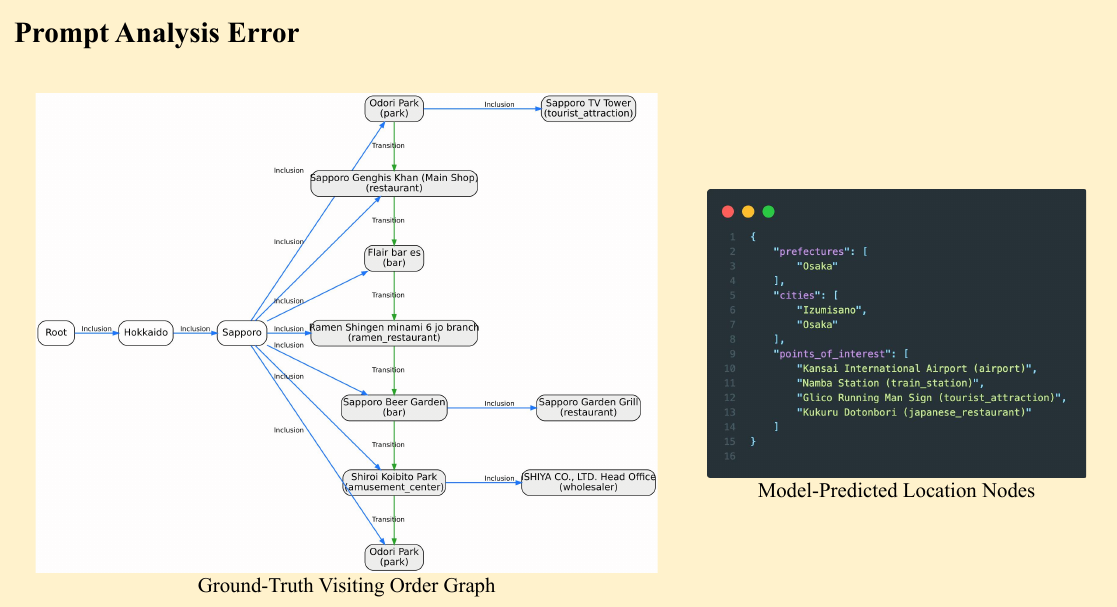}
    \caption{Example of a prompt analysis error made by the model during the node prediction task.}
    \label{fig:error-case-1}
\end{figure}

\paragraph{Node Prediction}
Figure~\ref{fig:error-case-1} is an output example from LLaVA-Video-7B for the video at \url{https://www.youtube.com/watch?v=2guZrrVMGfI}.
The model simply copied the example output provided in the input prompt (Table~\ref{tab:prompt:node-prediction}), without attempting to predict the actual visited locations in the video.
We believe this behavior is due to the relatively long prompt, which may have hindered the model’s ability to follow instructions effectively.

\begin{figure}[ht]
    \centering
    \includegraphics[width=0.97\linewidth]{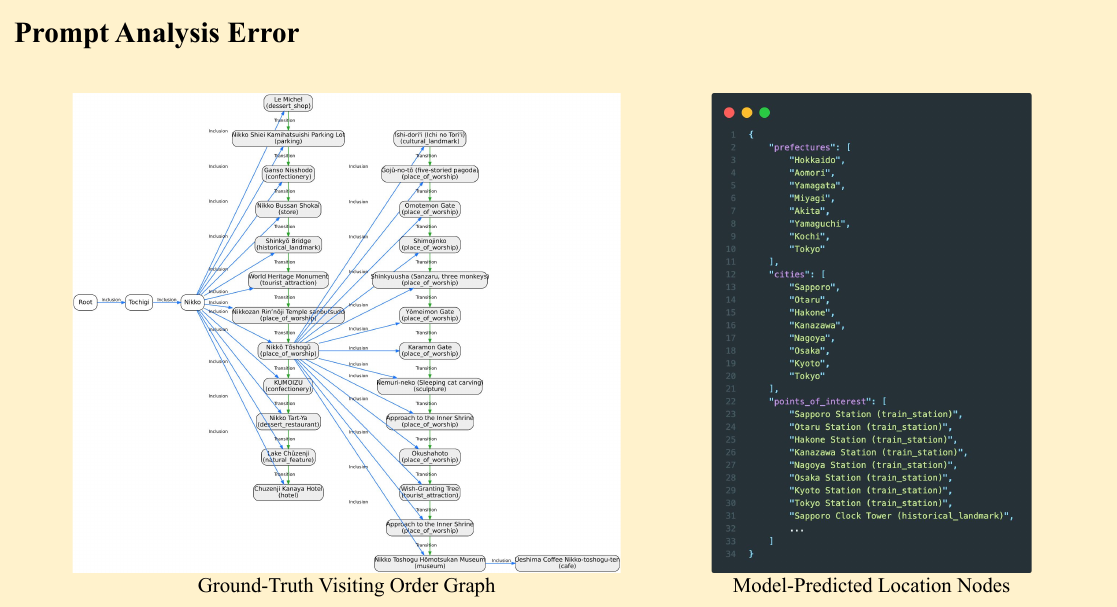}
    \caption{Another example of a prompt analysis error made by the model during the node prediction task.}
    \label{fig:error-case-2}
\end{figure}

Figure~\ref{fig:error-case-2} shows another example from LLaVA-Video-7B using the video at \url{https://www.youtube.com/watch?v=nzUKUdEHWAc}.
The video is a vlog filmed in \textit{Nikko, Tochigi}, but the model listed numerous prefectures, cities, and POIs, resembling random guessing.

\begin{figure}[ht]
    \centering
    \includegraphics[width=0.97\linewidth]{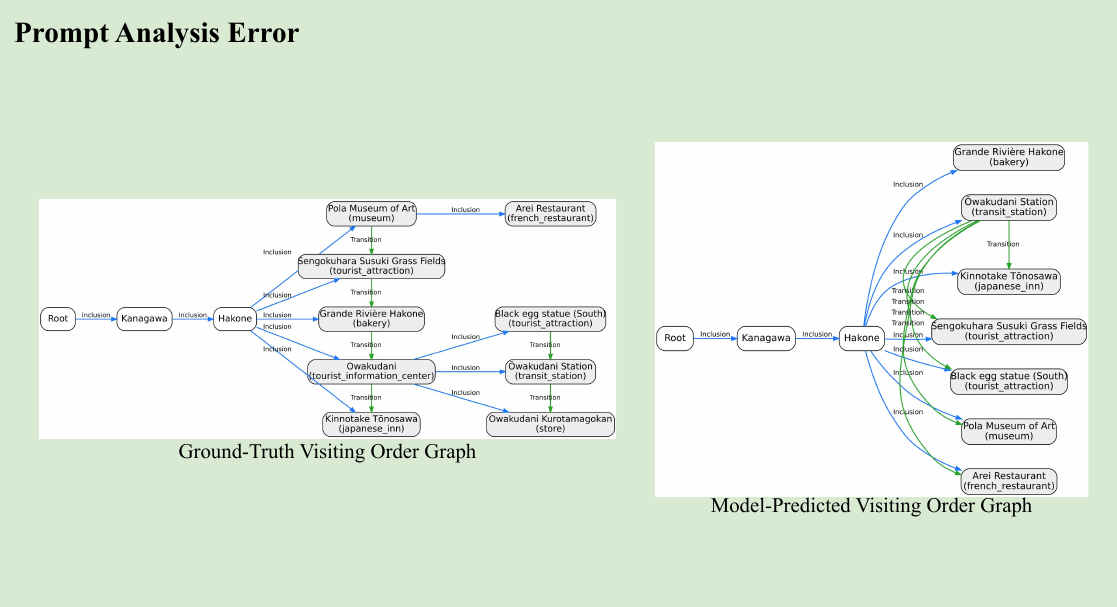}
    \caption{Example of a prompt analysis error made by the model during the edge prediction task.}
    \label{fig:error-case-3}
\end{figure}

\paragraph{Edge Prediction}
Figure~\ref{fig:error-case-3} presents an example from InternVL3-8B using the video at \url{https://www.youtube.com/watch?v=1YWj6ll-tJI}.
The model appeared to misinterpret the concept of a “sub-POI,” failing to capture the inclusion edge between \textit{Ōwakudani} and \textit{Ōwakudani Station}—a relation that should be identifiable even without geographic knowledge.
Additionally, its grasp of transition edges was insufficient: it generated five outgoing transition edges from \textit{Ōwakudani Station}, completely disregarding the video’s chronological order.

\subsubsection{Geographic Knowledge Error}
\begin{figure}[ht]
    \centering
    \includegraphics[width=0.97\linewidth]{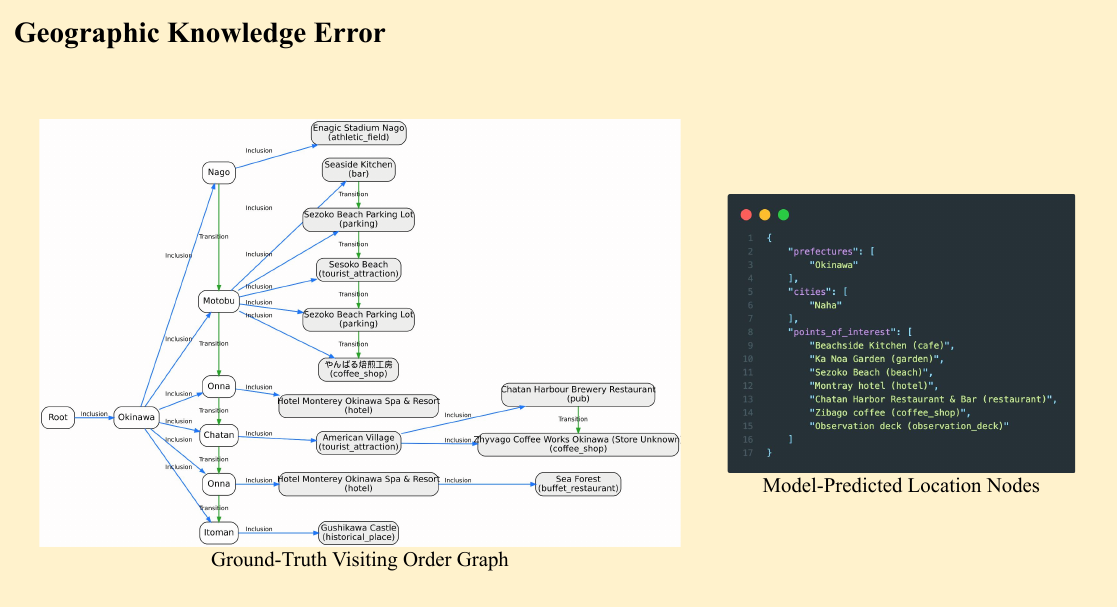}
    \caption{Example of a geographic knowledge error.}
    \label{fig:error-case-4}
\end{figure}

Figure~\ref{fig:error-case-4} shows an example from Qwen2.5-VL-7B using the video at \url{https://www.youtube.com/watch?v=x37h_iPtWiI}.
Although the model correctly identified the prefecture as \textit{Okinawa}, it failed to predict the correct cities, returning only \textit{Naha}, the most well-known city in Okinawa but not visited in the video.
Its POI predictions were also poor, with fewer than half of the locations correctly identified.
These results suggest that the model has a limited understanding of geographic knowledge.

\begin{figure}[ht]
    \centering
    \includegraphics[width=0.97\linewidth]{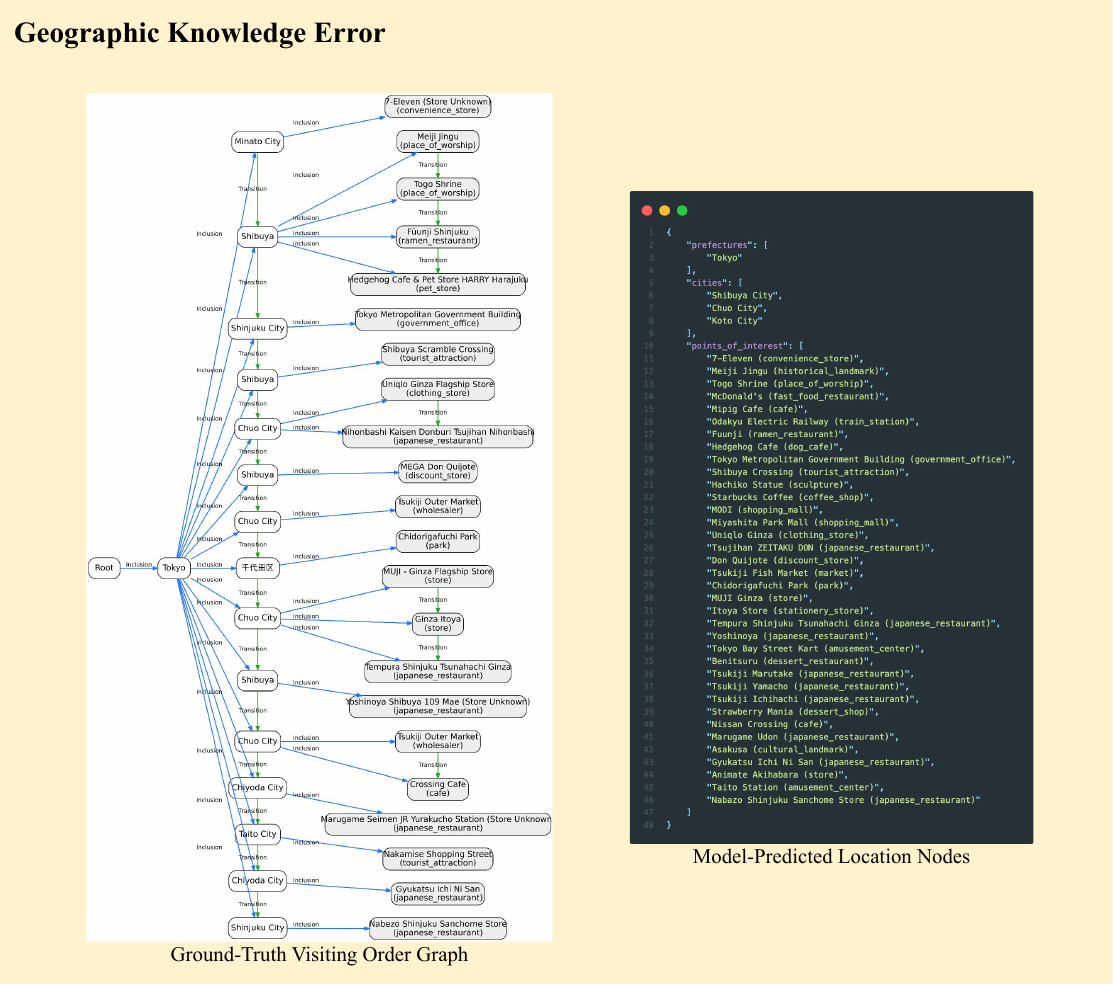}
    \caption{Another example of a geographic knowledge error.}
    \label{fig:error-case-5}
\end{figure}

Figure~\ref{fig:error-case-5} presents another example from Gemini-2.5-Flash using the video at \url{https://www.youtube.com/watch?v=RCpg1DHX-7s}.
While the predicted POIs are largely accurate, some cities featured in the video—such as \textit{Chiyoda City} and \textit{Shinjuku City}—were omitted.
Interestingly, the model correctly identified \textit{Nabazo Shinjuku Sanchome Store} but failed to associate it with \textit{Shinjuku City}, suggesting that even strong proprietary models still exhibit limitations in geographic reasoning.

\subsubsection{Temporal Reasoning Error}
\begin{figure}[ht]
    \centering
    \includegraphics[width=0.97\linewidth]{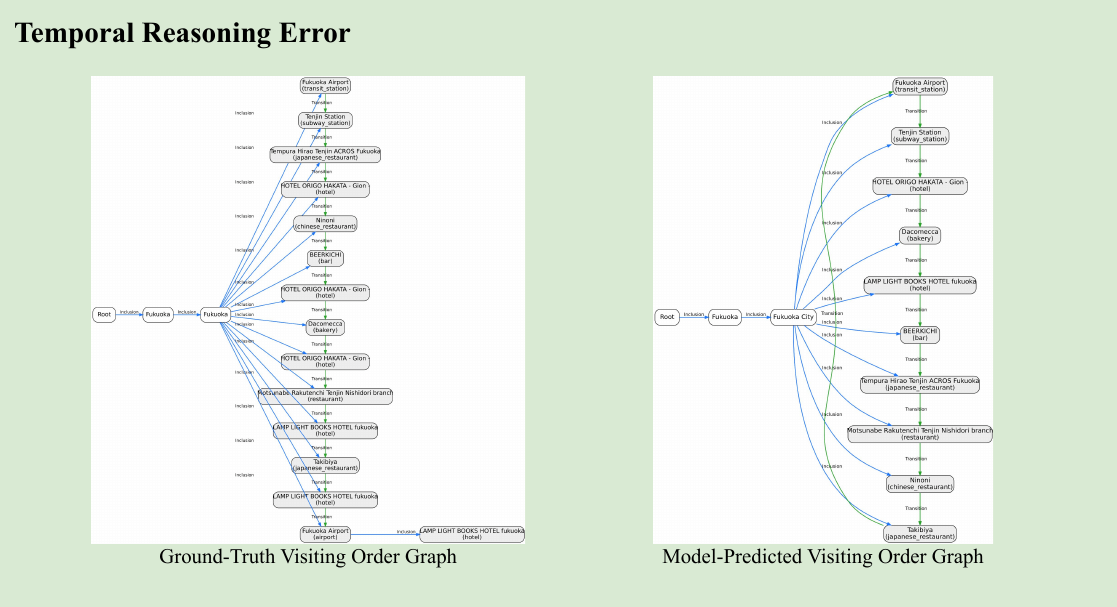}
    \caption{Example of a temporal reasoning error. Due to the output format, multiple visited locations in the model-predicted graph appear as a single node, unlike the multiple nodes in the ground-truth graph.}
    \label{fig:error-case-6}
\end{figure}

Figure~\ref{fig:error-case-6} presents an example from InternVL3-78B using the video at \url{https://www.youtube.com/watch?v=kBErL9FMce4}.
The video features multiple visited locations (e.g., \textit{HOTEL ORIGO HAKATA} and \textit{LAMP LIGHT BOOKS HOTEL}), but most of these visits are ignored by the model.
If predicted correctly, the model-predicted graph in Figure~\ref{fig:error-case-6} would contain many circles, as we visualize each visited location as a single node.
This suggests that the model failed to form a holistic view of the itinerary, instead constructing the graph in a piecemeal manner—adding edges one by one without considering the overall structure.

\begin{figure}[ht]
    \centering
    \includegraphics[width=0.97\linewidth]{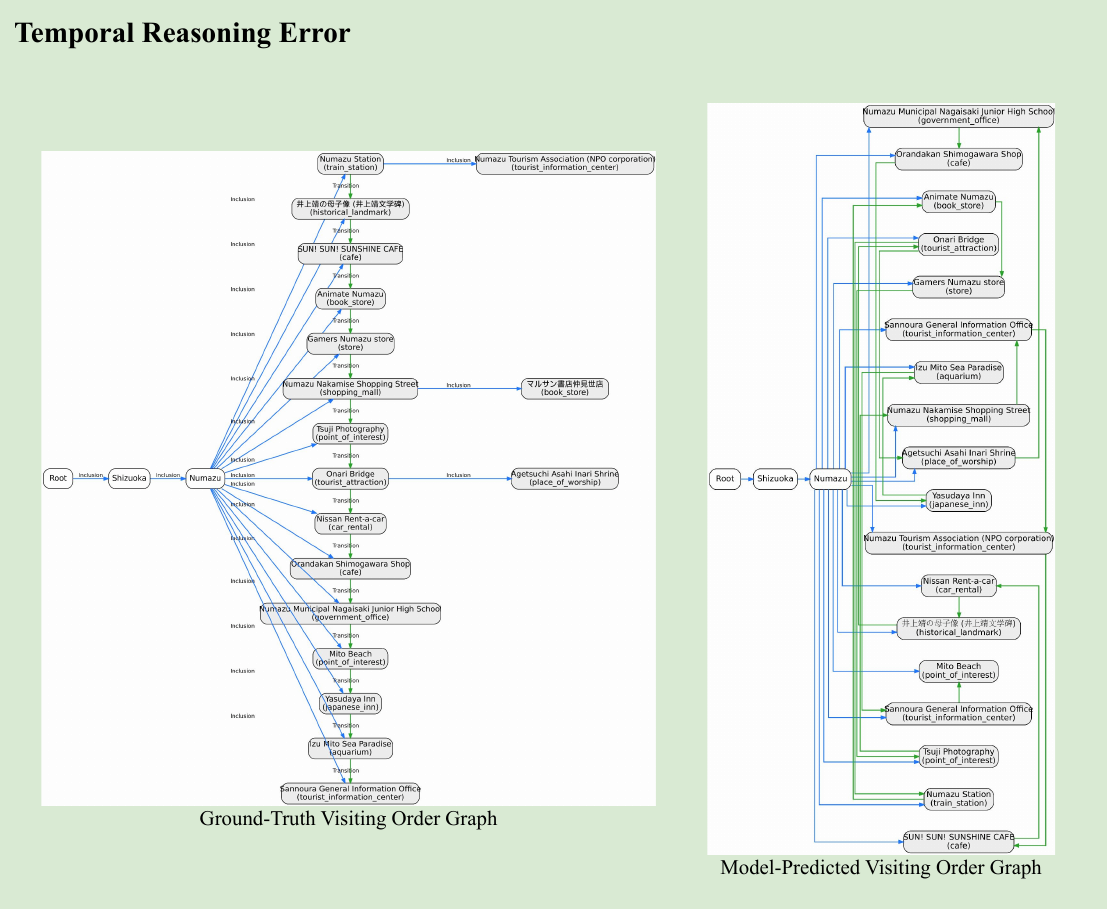}
    \caption{Another example of a temporal reasoning error.}
    \label{fig:error-case-7}
\end{figure}

Figure~\ref{fig:error-case-7} presents another example from o4-mini using the video at \url{https://www.youtube.com/watch?v=0Vjufc6rhH8}.
When the video contains a large number of POIs, the overall flow of the model-predicted graph becomes difficult to follow, indicating that even strong reasoning models like o4-mini still struggle with limited long-context and temporal reasoning capabilities.
While we attempted to obtain the thinking summary provided by OpenAI, the information was too sparse to yield meaningful insights (thinking summary for the generation process of Figure~\ref{fig:error-case-7} is shown in Table~\ref{tab:thinking-summary}).
We contend that the development of open-weight reasoning MLLMs is urgently needed, and VIR-Bench offers a challenging benchmark for evaluating their capabilities.

\begin{table*}[ht]
\centering
\small
\begin{tabular}{l}
\hline\\
\begin{minipage}{\columnwidth}
\begin{verbatim}
**Planning a visit to Numazu**
I want to explore Nakamise Shopping Street in Numazu. I’ll stroll through the street
and check out Marusan bookstore. Then, I’ll look for the Yasushi Inoue's Mother and
Child statue nearby. Next, I'll consider which landmark to visit — maybe Onari Bridge?
After that, I’ll visit Agetsuchi Asahi Inari Shrine, enjoying the river view. Finally,
I’ll head back to the station, catch the bus to Izu Mito Sea Paradise, and pass the
rental car location!
\end{verbatim}
\end{minipage}\\\\
\hline
\end{tabular}
\caption{Thinking summary of o4-mini during the generation process for Figure~\ref{fig:error-case-7}.}
\label{tab:thinking-summary}
\end{table*}
\clearpage
\subsection{Prompt Templates}
The prompts used for node and edge prediction are shown in Table~\ref{tab:prompt:node-prediction} and Table~\ref{tab:prompt:edge-prediction}, respectively.

\begin{table*}[ht]
\centering
\tiny
\begin{adjustwidth}{-2.2cm}{-2.2cm}
\begin{tabular}{l}
\begin{minipage}{\columnwidth}
\begin{verbatim}
---------------------------------------------------------------------------------------------------------------------------------------------------------------
### **Overview**

**Role**: You are an expert geo-location AI with a specialization in Japanese geography and travel analysis.

**Objective**: Analyze the provided video of a trip in Japan and identify all visited locations.

**Input**: A video file depicting travel within Japan.

### **Location Definitions**

1. **Prefecture (都道府県)**: The highest-level administrative division of Japan, such as Tokyo, Osaka, Hiroshima,Hokkaido, etc.

2. **City (東京23区、市町村)**: The municipality within a prefecture. This include Tokyo's 23 Specific Wards like Shibuya and Minato, cities like Yokohama and
Sapporo, towns like Karuizawa and Hakone, and villages like Hakuba and Shirakawa.

3. **Point of Interest (POI)**: A specific, named location. This includes landmarks, tourist attractions, stations, restaurants, cafes, stores, parks, museums,
etc.

### **Instructions & Rules**

1. **No Duplicates**: Each unique prefecture, city, and POI should appear only once in its respective list.
2. **Maximum Specificity**: Always use the most specific and complete name identifiable. For businesses, include the branch name if visible
(e.g., "Starbucks Shibuya Tsutaya" not just "Starbucks").
3. **Verification over Guesswork**: Only include locations that can be confidently identified from clear visual or auditory cues in the video
(e.g., signs, logos, building architecture, announcements). Do not guess or infer locations that are not explicitly evidenced.
4. **Minimum POI Unit**: Do not list integrated facilities within a POI (e.g., a museum's gift shop). However, you must list separate businesses located inside
a larger commercial complex, such as a specific café within a shopping mall or train station.
5. **Maximum POI Unit**: Do not include general geographic areas, districts, or neighborhoods which are not single, managed entities
(e.g., "Minato Mirai," "Kokusai Dori").
6. **POI Formatting**: Strictly adhere to the `"Name (category)"` format for all entries in the `points_of_interest` list. You must select the single most
appropriate category for each POI from the curated list below.

**POI Category List**:
['adventure_sports_center', 'airport', 'american_restaurant', 'amusement_center', 'amusement_park', 'aquarium', 'arena', 'art_gallery', ...

### **Output Format**

Generate a single JSON object conforming to the schema below. Ensure the lists are populated according to the rules above.

**JSON Schema**:

```json
{
  "prefectures": ["list of prefectures"],
  "cities": ["list of cities"],
  "points_of_interest": ["list of points of interest"]
}
```

**Example Output**:

For a video showing a person arriving at Kansai International Airport, taking a train to Osaka, seeing the Glico Running Man sign, and then eating at a Dotonbori
restaurant, the output should be:

```json
{
  "prefectures": [
    "Osaka"
  ],
  "cities": [
    "Izumisano",
    "Osaka"
  ],
  "points_of_interest": [
    "Kansai International Airport (airport)",
    "Namba Station (train_station)",
    "Glico Running Man Sign (tourist_attraction)",
    "Kukuru Dotonbori (japanese_restaurant)"
  ]
}
```
---------------------------------------------------------------------------------------------------------------------------------------------------------------
\end{verbatim}
\end{minipage}
\end{tabular}
\end{adjustwidth}
\caption{Prompt for node prediction.}
\label{tab:prompt:node-prediction}
\end{table*}

\begin{table*}[ht]
\centering
\tiny
\begin{adjustwidth}{-2.8cm}{-2.8cm}
\begin{tabular}{l}
\begin{minipage}{\columnwidth}
\begin{verbatim}
-------------------------------------------------------------------------------------------------------------------------------------------------------------------------
### **Overview**

**Role**: You are an expert geo-location AI with a specialization in Japanese geography and travel analysis.

**Objective**: Analyze the provided video of a trip in Japan. Based on the sequence of visited locations, generate a directed graph representing the visiting
trajectory. The graph should distinguish between hierarchical containment (e.g., a landmark within a city) and sequential travel (e.g., moving from one landmark
to another).

**Input**: A video file depicting travel within Japan and all visited locations.

### **Input**

The input for visited locations is a JSON object, as shown below. Crucially, the lists are **not** in chronological order.

```json
{{ input_json }}
```

### **Edge Definitions**

1. **Inclusion Edge**: A directed edge representing containment of one location within another. This edge flows from the larger geographical area to the smaller one.
    - Example: 
      - From a prefecture to a city: Osaka (prefecture) → Osaka (city)
      - From a city to a point of interest (POI): Shibuya → Shibuya Station
      - From a POI to a sub-POI: AEON MALL Zama → AEON Cinema Zama

2. **Transition Edge**: A directed edge representing movement between two distinct locations at the same hierarchical level. This edge indicates the chronological
flow of travel.
    - Example:
      - Between prefectures: Osaka → Hiroshima
      - Between cities (within the same prefecture): Shibuya → Shinjuku
      - Between POIs (within the same city): Tokyo Tower → Odaiba Marine Park

### **Instructions & Rules**

1. **Inclusion Edges**: 
   - For every city in the input, create an inclusion edge from its corresponding prefecture node.
   - For every POI in the input, create an inclusion edge from its corresponding city node.
   - For every sub-POI (if applicable) within a POI, create an inclusion edge from the POI node to the sub-POI node.
2. **Transition Edges**: 
   - Create transition edges between consecutively visited locations at the same hierarchical level.
   - A transition from a POI in City A to a POI in City B breaks the POI-level transition chain. The graph must instead show a transition edge from City A to City B.
   - Similarly, a transition from a city in Prefecture A to a city in Prefecture B is represented by a transition edge from Prefecture A to Prefecture B.
3. **Location Useage**:
    - Please make sure to use only the exact location names provided in the input—do not alter, add to, or infer any names beyond what is given.
4. **Unkonwn POIs**:
    - For POIs which can not be identified with certainty, we provide placeholders in the format `"Unknown (category)"` (e.g., Unknown (restaurant)). Use these
    placeholders when creating edges.

### **Output Format**

Generate a single JSON object conforming to the schema below. This object will represent the final directed graph with all nodes and edges correctly populated
according to the rules.

**JSON Schema**:

```json
{
    "edges": [
        {
        "source": "string",
        "target": "string",
        "type": "inclusion | transition"
        }
    ]
}
```

**Example Output**:

For a video showing a person arriving at Kansai International Airport, taking a train to Osaka, seeing the Glico Running Man sign, and then eating at a Dotonbori
restaurant, the output should be:

```json
{
    "edges": [
      {
        "source": "Osaka",
        "target": "Izumisano",
        "type": "inclusion"
      },
      ...
      {
        "source": "Glico Running Man Sign",
        "target": "Kukuru Dotonbori",
        "type": "transition"
      }
    ]
}
```
-------------------------------------------------------------------------------------------------------------------------------------------------------------------------
\end{verbatim}
\end{minipage}
\end{tabular}
\end{adjustwidth}
\caption{Prompt for edge prediction.}
\label{tab:prompt:edge-prediction}
\end{table*}\clearpage
\section{Travel Agent Details}
\subsection{Implementation Details}
\label{agent-implementation-details}

\begin{figure*}[ht]
  \centering
  \includegraphics[width=\linewidth]{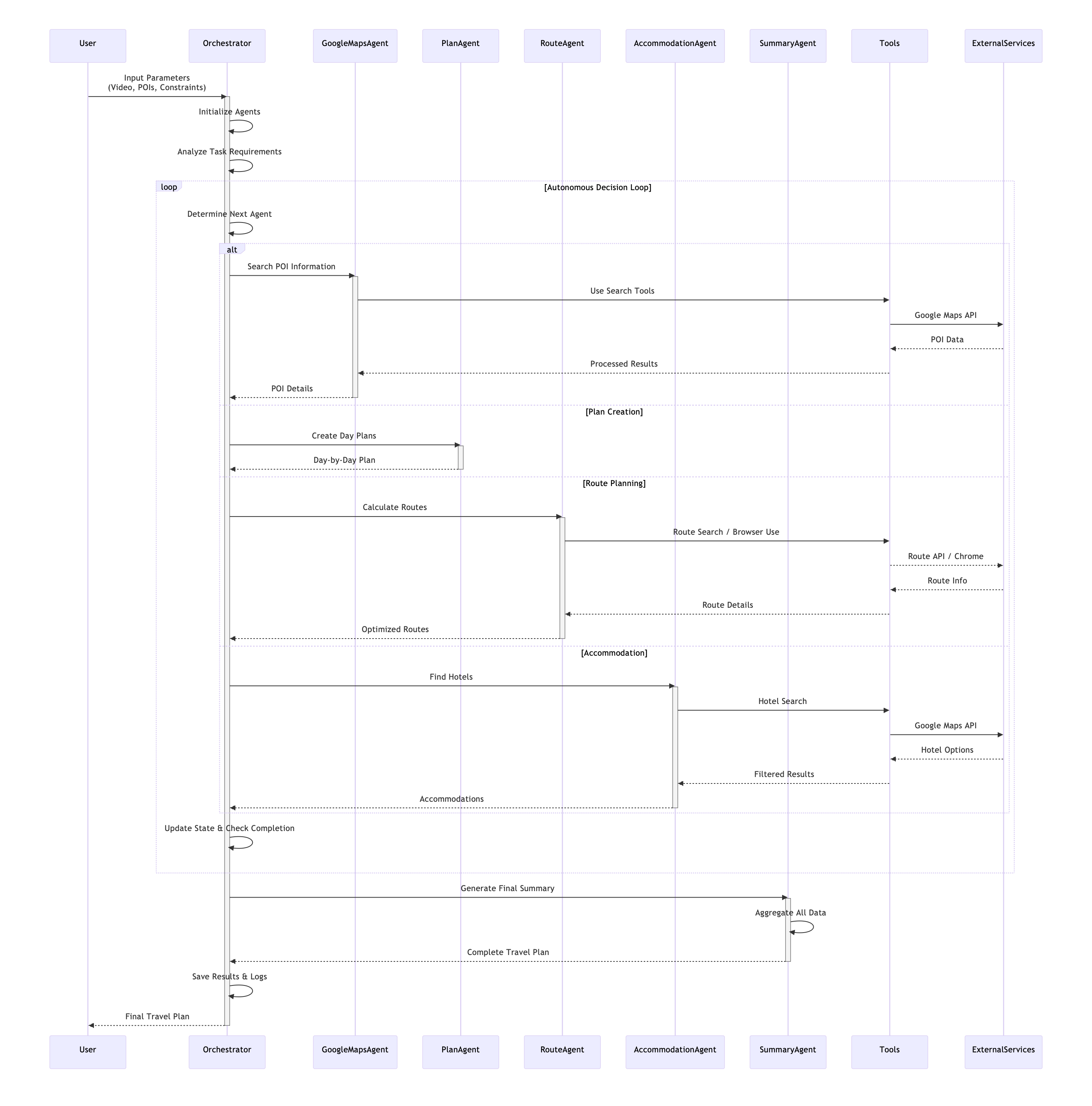}
  \caption{The sequence diagram of the agent system.}
  \label{fig:agent-flow}
\end{figure*}

We implemented the travel plan agent system using LangChain\footnote{\url{https://www.langchain.com}}.
The system consists of one orchestrator, five specific agents and some tools.
The sequence diagram of this system is shown in Figure~\ref{fig:agent-flow}.

As tools, following external services are used:
\begin{itemize}
    \item Google Places API: For detailed POI information and accommodation information.
    \item Google Routes API: For route search (``TRANSIT'' mode is unavailable in Japan).
    \item BrowserUse\footnote{\url{https://github.com/browser-use}}: For fallback of the route search by automated GUI-based browser operation.
\end{itemize}

We randomly set the input constraints as below:
\begin{itemize}
    \item Headcount: 1 to 4
    \item Duration: 2 days to 4 days.
    \item Budget per day per person: 70 USD to 500 USD.
\end{itemize}

\subsection{Evaluation Details}

\begin{figure*}[t]
  \centering
  \subfloat[Attraction Task]{
    \includegraphics[width=0.45\textwidth, valign=b]{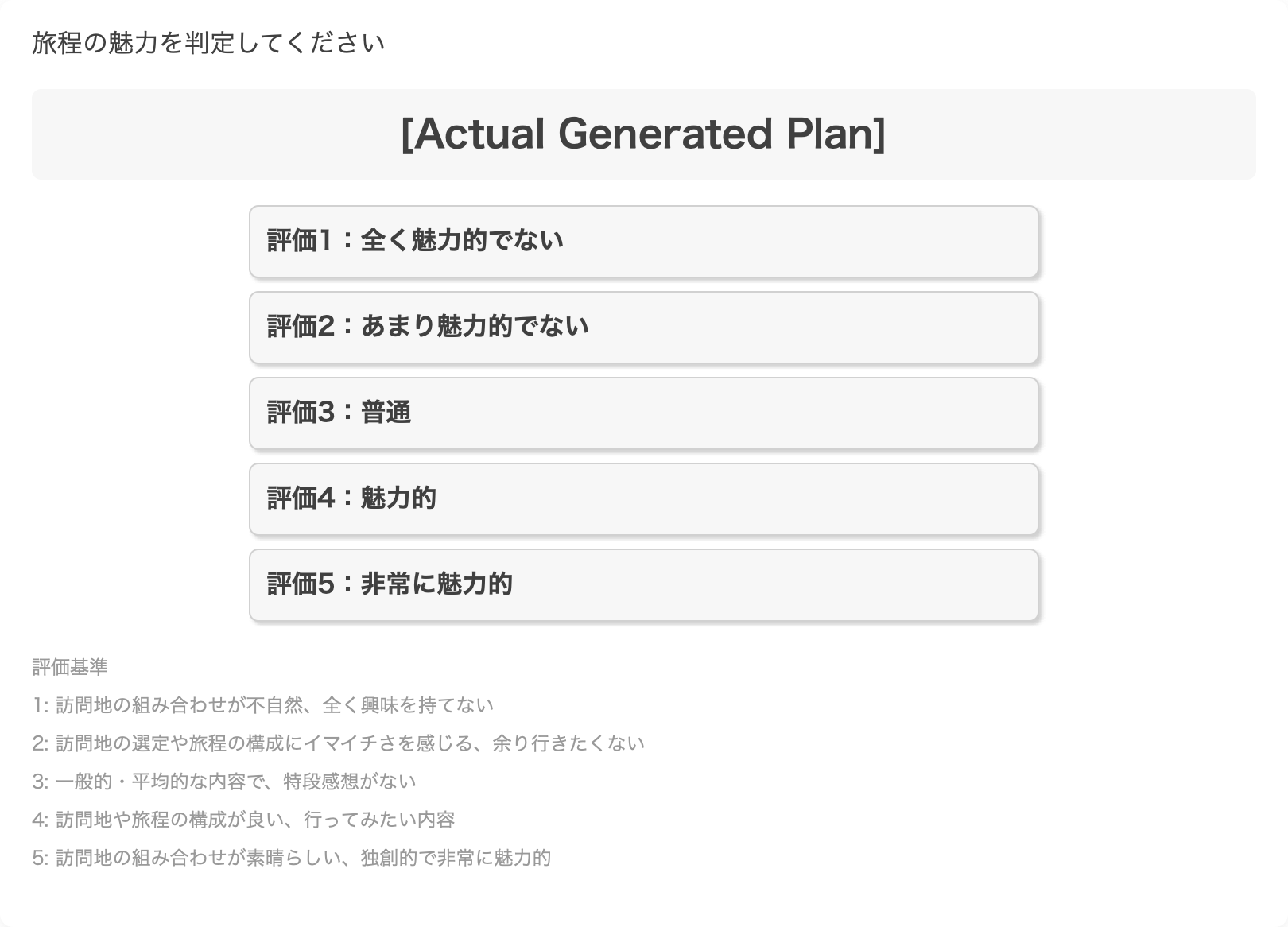}
    \label{fig:cs_attract_apdx}
  }
  \subfloat[Feasibility Task]{
    \includegraphics[width=0.45\textwidth, valign=b]{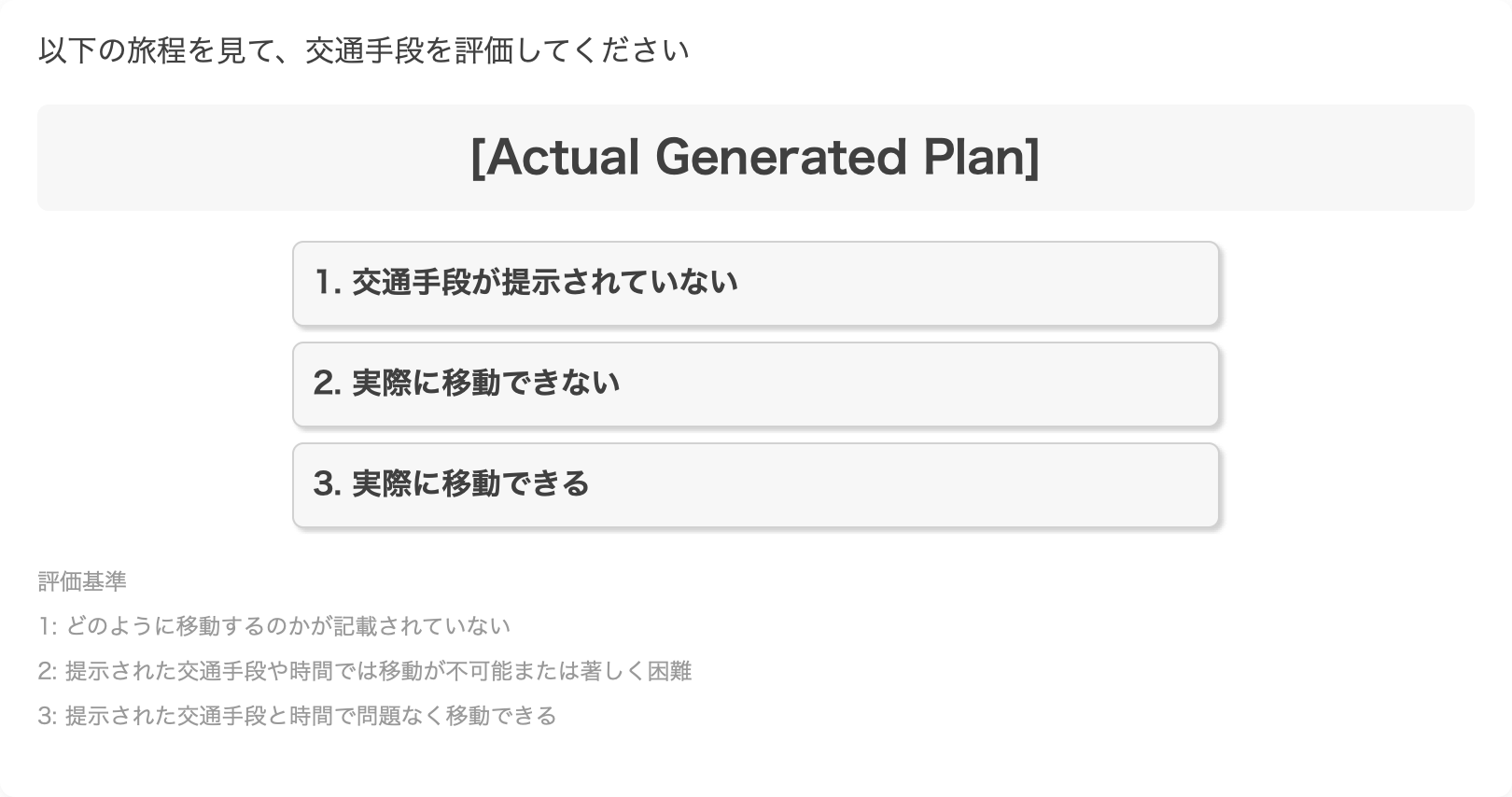}
    \label{fig:cs_transport_apdx}
  }
  \\
  \subfloat[Density Task]{
    \includegraphics[width=0.45\textwidth, valign=b]{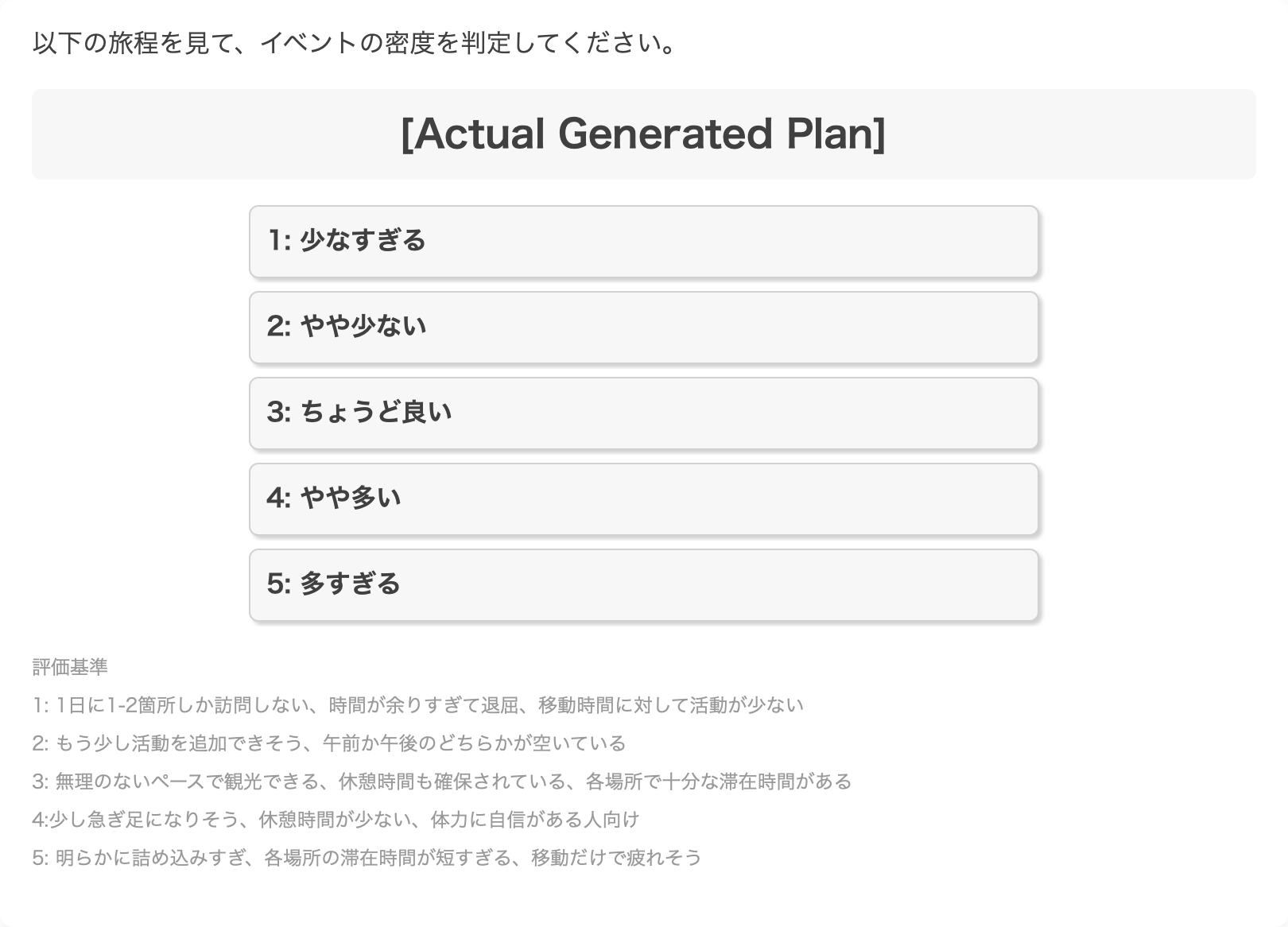}
    \label{fig:cs_density_apdx}
  }
  \subfloat[Alignment Task]{
    \includegraphics[width=0.45\textwidth, valign=b]{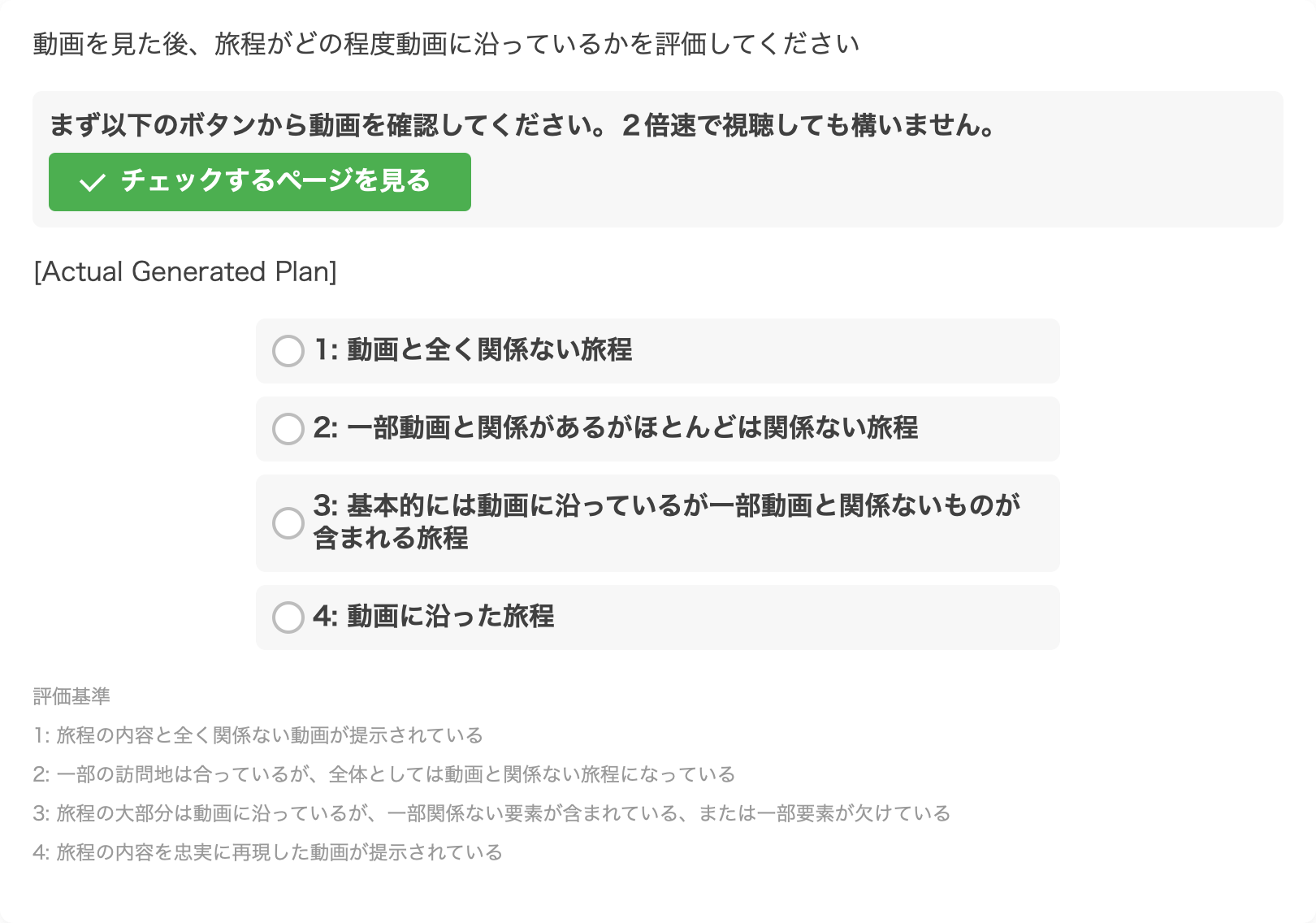}
    \label{fig:cs_relevance_apdx}
  }
  \caption{Crowdsourcing screen examples. ``[Actual Generated Plan]'' is substituted with the actual generated plans.}
  \label{fig:cs_screen}
\end{figure*}

We have evaluated generated plans using crowdsourcing on Yahoo! Crowdsourcing\footnote{\url{https://crowdsourcing.yahoo.co.jp}}.
The instructions to crowdworkers and the screen where the workers execute the tasks are shown in Table~\ref{tab:instruction:attract},~\ref{tab:instruction:trans},~\ref{tab:instruction:density},~\ref{tab:instruction:relevance} and Figure~\ref{fig:cs_screen}, respectively.
The shown instructions are translated into English while they are originally in Japanese.

To ensure the quality of the evaluation, we included check questions with obvious correct answers and only accepted responses from users who answered them correctly.
One check question was assigned to each worker.
In the alignment task, we asked the workers to watch videos and then answer the questions, as shown in Figure~\ref{fig:cs_relevance}.

The detailed crowdsourcing results corresponding to Figure 4 in the main paper are shown in Table~\ref{tab:aggregated_analysis}.

\begin{table*}[htbp]
  \centering
  \resizebox{\textwidth}{!}{
  \begin{tabular}{l|ccccccccccccccccc}
    \toprule
    & \multicolumn{5}{c}{Attraction} & \multicolumn{5}{c}{Density} & \multicolumn{3}{c}{Feasibility} & \multicolumn{4}{c}{Alignment} \\
    \cmidrule(lr){2-6} \cmidrule(lr){7-11} \cmidrule(lr){12-14} \cmidrule(lr){15-18}
    Setting & Very Attr. & Attr. & Neutral & Not Attr. & Not Attr. At All & Too Many & Slightly Many & Just Right & Slightly Few & Too Few & Navigable & Not Navigable & No Info & Matches & Mostly Matches & Partially Matches & No Match \\
    \midrule
    POI & 18.0 & 41.0 & 28.0 & 7.0 & 6.0 & 2.0 & 24.0 & 48.0 & 20.0 & 6.0 & 82.0 & 11.0 & 7.0 & 10.0 & 25.0 & 54.0 & 11.0 \\
    Video & 17.0 & 35.0 & 32.0 & 9.0 & 7.0 & 3.0 & 27.0 & 45.0 & 24.0 & 1.0 & 70.0 & 10.0 & 20.0 & 12.0 & 29.0 & 28.0 & 31.0 \\
    POI + Video & 20.0 & 47.0 & 22.0 & 8.0 & 3.0 & 4.0 & 26.0 & 53.0 & 13.0 & 4.0 & 78.0 & 11.0 & 11.0 & 3.0 & 24.0 & 51.0 & 22.0 \\
    \bottomrule
  \end{tabular}}
  \caption{Aggregated analysis of user evaluation ratings across different settings (all values in \%)}
  \label{tab:aggregated_analysis}
\end{table*}

\subsection{Examples of the Generated Plans}
Actual plans generated for each agent setting are shown in Sections~\ref{ssection:plan_poi}, \ref{ssection:plan_video}, and \ref{ssection:plan_both}.
The input video used for these examples is: \url{https://www.youtube.com/watch?v=P60ee5-wQic}.
For visibility, the plans were converted from Markdown to LaTeX using a rule-based script.

The generated plans highlight the appeal of POIs and provide practical information such as prices and transportation options.
The input video is filmed in Kyoto and Nara, however, Section~\ref{ssection:plan_video} contains a plan in Tokyo.
As mentioned in the main paper, the video-only setting sometimes generates completely unrelated itineraries.

\clearpage\begin{table*}[ht]
\centering
\tiny
\begin{adjustwidth}{-2.2cm}{-2.2cm}
\begin{tabular}{l}
\begin{minipage}{\columnwidth}
\begin{verbatim}
---------------------------------------------------------------------------------------------------------------------------------------------------------------
・ Travel Plan Appeal Evaluation Task

Task Overview

Read the travel plans generated by AI and evaluate them on a 5-point scale based on whether they are appealing enough to make you want to visit.

Evaluation Criteria

Evaluate the overall appeal of the plan, taking into consideration the appeal of the destinations and the specificity of the descriptions.

Rating 5: Very appealing

Criteria: The ideal plan. The combination of destinations is excellent, and the plan is full of specific and appealing information, such as“Let's try the
local specialty 〇〇.” It is original, and you feel strongly that you want to travel according to this plan.

Rating 4: Attractive

Criteria: A well-made, good plan. The destinations are attractive, and the itinerary flows smoothly. The information is specific and sufficient, and many
people will be satisfied with the content.

Rating 3: Average

Criteria: Gives the impression of a list of typical tourist destinations. Not bad, but nothing special. The information is mostly abstract, such as “visit XX,”
and the content is commonplace.

Rating 2: Somewhat unattractive

Criteria: Not a very appealing plan. There are questions about the choice of places to visit, and the information is not specific enough. It is difficult to
imagine how enjoyable the trip will be.

Rating 1: Not attractive at all

Criteria: It is difficult to understand why this plan was chosen. The combination of places to visit is incongruous, and there is very little information,
making it completely unattractive.

Points to note when evaluating

Attractiveness of destinations: Are the selected locations attractive as travel themes or destinations? Are there any suggestions other than the usual
tourist spots, such as hidden gems?

Specificity of information: The appeal can vary greatly depending on whether it simply says “lunch” or “fresh seafood bowl lunch at XX market.”

Plan structure: Is the theme of the trip (e.g., gourmet, scenic views, relaxation) consistent, and does it flow smoothly?

Excitement: Does reading the plan overall make you feel like “This sounds fun!” or “I want to go!”?

Response Method

Read the entire travel plan and select the option that best applies from the choices below. If you are unsure, consider whether you would choose this plan if
you were spending your own money and time.
---------------------------------------------------------------------------------------------------------------------------------------------------------------
\end{verbatim}
\end{minipage}
\end{tabular}
\end{adjustwidth}
\caption{The crowdsourcing instruction for the attraction task.}
\label{tab:instruction:attract}
\end{table*}\clearpage

\begin{table*}[ht]
\centering
\tiny
\begin{adjustwidth}{-2.2cm}{-2.2cm}
\begin{tabular}{l}
\begin{minipage}{\columnwidth}
\begin{verbatim}
---------------------------------------------------------------------------------------------------------------------------------------------------------------
・ Evaluate the appropriateness of transportation methods in the itinerary
Task overview

Read the travel plan generated by AI and evaluate whether the means of transportation used to travel between locations are realistic and appropriate.

Evaluation criteria

Check whether the travel times and means of transportation listed in the plan match the actual geography and traffic conditions, and select the most
appropriate option from the following three choices.

1. No transportation means specified

Criteria: No information is provided on how to travel (e.g., walking, train, bus, car).

Specific example:

“[10:00] Tokyo Station → [11:00] Senso-ji Temple”—the method of transportation to the next destination is not specified.

2. Travel is not possible

Criteria: The transportation method or travel time provided is unrealistic, making travel impossible or extremely difficult.
Specific example:

Travel time is too short: “30 minutes from Tokyo Station to Osaka Station by Shinkansen”

Inappropriate transportation method: “Walk from Okinawa Main Island to Miyakojima”

Physically impossible: “5 minutes on foot from Shibuya to Skytree”

3. Actual travel is possible

Criteria: The transportation method and travel time provided are realistic and travel is considered possible without undue difficulty.

Specific examples:

“【9:00】Shinjuku Station → (approx. 7 minutes by JR Yamanote Line) → 【9:10 approx.】Harajuku Station”

“[1:00 p.m.] Kinkaku-ji Temple → (approx. 40 minutes by city bus) → [approx. 1:40 p.m.] Kiyomizu-dera Temple”
Evaluation points

Imagine the map: Get a rough idea of the distance between locations.

Reasonable travel time: Consider waiting times for trains and buses, as well as transfer times, and determine whether the travel time provided is
feasible.

Regional characteristics: The appropriateness of transportation methods may vary depending on whether the area is an urban area (with well-developed
transportation networks) or a rural area (where a car is essential, etc.).

Answering method

Read the transportation sections of each travel plan and select the most appropriate option from 1 to 3. If you are unsure, consider whether you would choose
this transportation method or think it is feasible within the given time frame.
---------------------------------------------------------------------------------------------------------------------------------------------------------------
\end{verbatim}
\end{minipage}
\end{tabular}
\end{adjustwidth}
\caption{The crowdsourcing instruction for the feasibility task.}
\label{tab:instruction:trans}
\end{table*}\clearpage

\begin{table*}[ht]
\centering
\tiny
\begin{adjustwidth}{-2.2cm}{-2.2cm}
\begin{tabular}{l}
\begin{minipage}{\columnwidth}
\begin{verbatim}
---------------------------------------------------------------------------------------------------------------------------------------------------------------
・ Travel Plan Event Density Evaluation Task

Task Overview

Read the travel plan generated by AI and evaluate whether the number of activities per day (event density) is appropriate on a 5-point scale.

Evaluation Criteria

Event density: The degree to which sightseeing, meals, transportation, and activities are packed into a day.

Evaluation 1: Too few

Criteria: Only 1-2 locations visited per day, too much free time, not enough activities relative to travel time

Rating 2: Slightly insufficient

Criteria: Could use a few more activities, either the morning or afternoon is too empty

Rating 3: Just right

Criteria: Reasonable pace for sightseeing, adequate rest time, sufficient time at each location

Rating 4: Slightly excessive

Criteria: May feel rushed, limited rest time, suitable for those with good physical stamina

Rating 5: Too much

Criteria: Clearly overloaded, insufficient time at each location, may be tiring due to travel alone

Notes when evaluating

Consider the number of travelers and the number of days (specified at the beginning of the plan)

Include travel time as part of the activities

Ensure that meal times are appropriately allocated

Evaluate based on whether it is realistically feasible.

Evaluation examples

Example of Rating 1: “A two-day trip with hotel check-in only in the afternoon on the first day, and one sightseeing spot in the morning on the second day.”

Example of Rating 3: “Departure at 9 AM, visit 2-3 sightseeing spots at a reasonable pace, with time for lunch and dinner, and return to the hotel around 8 PM.”

Example of Evaluation 5: “Depart at 7 AM, visit 10 or more tourist spots, stay at each location for about 15 minutes, eat meals while traveling, and schedule
until 11 PM.”

Answering Method

Read each travel plan and select a number from 1 to 5. If you are unsure, consider whether you would actually want to travel according to the plan.
---------------------------------------------------------------------------------------------------------------------------------------------------------------
\end{verbatim}
\end{minipage}
\end{tabular}
\end{adjustwidth}
\caption{The crowdsourcing instruction for the density task.}
\label{tab:instruction:density}
\end{table*}\clearpage

\begin{table*}[ht]
\centering
\tiny
\begin{adjustwidth}{-2.2cm}{-2.2cm}
\begin{tabular}{l}
\begin{minipage}{\columnwidth}
\begin{verbatim}
---------------------------------------------------------------------------------------------------------------------------------------------------------------
・ Watch the vlog and determine whether the itinerary has been accurately reproduced.

Task Overview

Watch the travel vlog (video blog) and evaluate on a 4-point scale whether the itinerary generated by AI accurately reproduces the content of the video.
You may watch the video at double speed.

Evaluation Criteria

Check how accurately the places visited, meals, activities, etc. introduced in the video are reflected in the itinerary, and select the most appropriate option
from the following four choices.

1. The itinerary is completely unrelated to the video

Criteria: The places and activities listed in the itinerary are completely different from the content of the video.

Specific example:

The video is about a trip to Kyoto, but the itinerary is for a trip to Okinawa.

None of the tourist spots featured in the video are included in the itinerary.

2. The itinerary is partially related to the video but mostly unrelated

Criteria: The itinerary includes one or two locations featured in the video, but overall it is a completely different itinerary.

Specific example:

Only “Kiyomizu Temple,” which appeared in the Kyoto trip video, is included in the itinerary, but the other destinations and order are completely different.

3. The itinerary is basically the same as the video, but includes some parts that are not related to the video
Criteria: Most of the itinerary follows the content of the video, but there are additional places that are not shown in the video,
or parts of the video are missing.

Specific example:

The order of visits in the video (A→B→C) is reproduced, but a place called “D” that is not in the video has been added to the itinerary.

The itinerary covers all the major tourist spots visited in the video, but a café that was visited is missing.

4. Itinerary that follows the video

Criteria: The itinerary faithfully reproduces the content of the video, including the places visited, the order of visits, and the meals.

Specific example:

The sequence introduced in the video (“Breakfast at A Café → Sightseeing at B Temple → Street food at C Shopping Street”) is listed exactly as it appears
in the itinerary.

Points to note when evaluating

Order of visits: It is important that the places visited are correct. Minor differences in order are not a major issue.

Check that the specific names of shops and dishes mentioned in the video are reflected in the itinerary.

There is no need to recreate every action in the video (e.g., stopping at a convenience store, getting slightly lost). As long as the main events are covered,
it is acceptable.

Answering Method

Compare the vlog and itinerary, then select the option that best matches from 1 to 4. If you are unsure, use the following criterion: “Can you relive the trip
from the video by following this itinerary?”
---------------------------------------------------------------------------------------------------------------------------------------------------------------
\end{verbatim}
\end{minipage}
\end{tabular}
\end{adjustwidth}

\caption{The crowdsourcing instruction for alignment task.}
\label{tab:instruction:relevance}
\end{table*}\clearpage

\clearpage

\subsubsection{Generated Plan by POI-only Setting}
\tiny
\label{ssection:plan_poi}
\input{tex_plans/poi}
\clearpage

\subsubsection{Generated Plan by video-only Setting}
\label{ssection:plan_video}
\input{tex_plans/video}
\clearpage

\subsubsection{Generated Plan by P+V Setting}
\label{ssection:plan_both}
\input{tex_plans/both}

%% file: tex_plans/poi.tex
{\Large\bfseries\color{MidnightBlue} Travel Guide}\par\medskip

\noindent
\textbf{Destination}: N/A \\
\textbf{Travel Dates}: 2 days \\
\textbf{Group Size}: 2 people \\
\textbf{Budget}: \$1964.00 USD

\hrulefill

\section*{Your Unforgettable 2-Day Kyoto \& Nara Adventure}

Welcome to a whirlwind journey through ancient traditions and serene nature! This itinerary is crafted to give you and your travel partner a deep, immersive experience, balancing iconic sights with authentic local flavors. We've based your adventure in Kyoto, the cultural soul of Japan, with a magical day trip to Nara to meet its famous, friendly deer. Get ready to create some incredible memories together.

This plan focuses on Kyoto and Nara because, with just two days, it allows for a rich and relaxed experience without the rush of long-distance travel. The incredible sights of Tokyo, like Shibuya Crossing and Sensō-ji Temple, are amazing but would require a separate trip to be fully appreciated. We've prioritized depth over distance to make your short trip truly special.

\hrulefill

\section*{Day 1: Kyoto's Spiritual Peaks \& Culinary Heart}

\textbf{Theme:} A day of contrasts, from a spiritual mountain hike through thousands of vermilion gates to a sensory explosion in Kyoto's most famous food market.

\subsection*{Morning (8:00 AM – 12:00 PM): Fushimi Inari Taisha}

\begin{itemize}[leftmargin=*, nosep]
  \item \textbf{Why it's a must-see:} With an importance score of 91.8 and over 79,000 reviews, Fushimi Inari is an unmissable Kyoto icon, famous for the \textbf{Senbon Torii}, a mesmerizing network of thousands of vibrant red gates.
  \item \textbf{What to Expect:} Dedicated to Inari, the Shinto god of rice and business. The path is a hike, not just a stroll. The lower sections are crowded, but the upper trails are peaceful and atmospheric.
  \item \textbf{Recommended Duration:} 2.5–3 hours.
  \item \textbf{Optimal Timing:} Arrive by 8:00 AM for fewer crowds and stunning photos.
  \item \textbf{Special Considerations:} Comfortable walking shoes are essential. Consider turning around at the Yotsutsuji intersection if pressed for time.
\end{itemize}

\subsection*{Lunch (12:30 PM – 1:30 PM): Authentic Udon Near the Shrine}

\begin{itemize}[leftmargin=*, nosep]
  \item \textbf{Restaurant:} Kendonya
  \item \textbf{Address:} 25-13 Fukakusa Ichinotsubocho, Fushimi Ward, Kyoto
  \item \textbf{Cuisine:} Udon Noodles
  \item \textbf{Price Range:} Budget-friendly (\$)
  \item \textbf{Why it Complements the Itinerary:} Close to the shrine, this cozy spot serves handmade udon and is loved by locals.
  \item \textbf{Reservation:} Not required, but expect a short wait during peak times.
\end{itemize}

\subsection*{Afternoon (2:00 PM – 5:00 PM): Nishiki Market \& Teramachi Arcade}

\begin{itemize}[leftmargin=*, nosep]
  \item \textbf{Transportation:} Take the Keihan Main Line to Gion-Shijo Station (\~10 mins, \$2 USD), then walk 5 minutes to the market.
  \item \textbf{Why it's a must-see:} Known as "Kyoto’s Kitchen", Nishiki Market is a five-block arcade packed with food stalls, traditional goods, and vibrant atmosphere.
  \item \textbf{What to Expect:} Sample everything from candied octopus to fresh sashimi. Explore nearby Teramachi for modern boutiques and souvenirs.
  \item \textbf{Recommended Duration:} 2–3 hours.
\end{itemize}

\subsection*{Dinner (6:30 PM): A Taste of Modern Kyoto}

\begin{itemize}[leftmargin=*, nosep]
  \item \textbf{Restaurant:} Cross Hotel Kyoto Restaurant \& Bar
  \item \textbf{Cuisine:} Japanese-Italian Fusion
  \item \textbf{Price Range:} Moderate-High (\$\$\$)
  \item \textbf{Why it Complements the Itinerary:} Conveniently located in your hotel, it offers a stylish yet relaxing end to the day.
  \item \textbf{Reservation:} Recommended; ask the hotel concierge.
\end{itemize}

\hrulefill

\section*{Day 2: A Day with the Deer in Ancient Nara}

\textbf{Theme:} A delightful and easy day trip to Japan's first permanent capital, where sacred deer roam among ancient temples and tranquil paths.

\subsection*{Morning (9:00 AM – 1:00 PM): Journey to Nara \& Todai-ji Temple}

\begin{itemize}[leftmargin=*, nosep]
  \item \textbf{Transportation:} JR Nara Line Rapid train from Kyoto Station to JR Nara Station (\~45 mins, \$6 USD one-way).
  \item \textbf{Activity: Nara Park:} Meet the friendly bowing deer, roam the park, and enjoy nature.
  \item \textbf{Explore Todai-ji:} Visit the Great Buddha inside the world’s largest wooden structure.
  \item \textbf{Cost:} Nara Park is free; Todai-ji Temple: ¥600 (\~\$5 USD).
\end{itemize}

\subsection*{Lunch (1:00 PM – 2:00 PM): Award-Winning Burgers}

\begin{itemize}[leftmargin=*, nosep]
  \item \textbf{Restaurant:} SAKURA BURGER
  \item \textbf{Address:} 38-1 Higashimuki Kitamachi, Nara
  \item \textbf{Cuisine:} American Gourmet Burgers
  \item \textbf{Price Range:} Moderate (\$\$)
  \item \textbf{Why it Complements the Itinerary:} Highly-rated and a nice change from traditional Japanese meals.
\end{itemize}

\subsection*{Afternoon (2:00 PM – 4:30 PM): Serene Shrines \& Gardens}

\begin{itemize}[leftmargin=*, nosep]
  \item \textbf{Explore:} Stroll to Kasuga Taisha Shrine through a tranquil lantern-lined forest path.
  \item \textbf{Return:} Head back to JR Nara Station and return to Kyoto.
\end{itemize}

\subsection*{Dinner (7:00 PM): Crispy Tonkatsu in Nara or Kyoto}

\begin{itemize}[leftmargin=*, nosep]
  \item \textbf{Restaurant:} Marukatsu Nara Honten (or nearby Kyoto options)
  \item \textbf{Cuisine:} Tonkatsu (Pork Cutlet)
  \item \textbf{Price Range:} Budget-friendly (\$)
  \item \textbf{Why it Complements the Itinerary:} A hearty and satisfying way to end the day before returning to your hotel.
\end{itemize}

\hrulefill

\section*{Accommodation Recommendation}

\textbf{Hotel:} Cross Hotel Kyoto \\
\textbf{Address:} 71-1 Daikokuchō, Nakagyo Ward, Kyoto \\
\textbf{Rating:} 4.6 stars (1,796 reviews)

\begin{itemize}[leftmargin=*, nosep]
  \item \textbf{Why it's Perfect for You:} Combines comfort, design, and prime location. Great for couples, near Nishiki Market and Gion.
  \item \textbf{Recommended Room:} Standard or Superior King Room
\end{itemize}

\hrulefill

\section*{Budget Breakdown (for 2 People)}

\begin{itemize}[leftmargin=*, nosep]
  \item \textbf{Total Budget:} \$1964.00 USD
  \item \textbf{Estimated Total Cost:} \$1964.00 USD
  \item \textbf{Utilization:} 100\%
\end{itemize}

\textbf{Detailed Breakdown:}
\begin{itemize}[leftmargin=*, nosep]
  \item Accommodation: \$N/A
  \item Transportation: \$N/A
  \item Activities: \$N/A
  \item Food: \$N/A
\end{itemize}

\hrulefill

\section*{Practical Travel Tips for Two}

\begin{itemize}[leftmargin=*, nosep]
  \item \textbf{Get an IC Card:} Use Suica or ICOCA for effortless train and bus travel.
  \item \textbf{Share and Sample:} Share snacks at Nishiki Market to try more without getting full.
  \item \textbf{Pack Smart:} Layer clothing and wear comfortable shoes.
  \item \textbf{Cash is Still King:} Many local shops and food stalls only accept yen.
  \item \textbf{Embrace the Quiet:} Respectful silence in shrines enhances the experience.
\end{itemize}

\vspace{1em}
\begin{center}
    \Large\bfseries\color{MidnightBlue} Enjoy every moment of your incredible Japanese adventure!
\end{center}

%% file: tex_plans/video.tex
{\Large\bfseries\color{MidnightBlue} Travel Guide}\par\medskip

\noindent
\textbf{Destination}: N/A \\
\textbf{Travel Dates}: 2 days \\
\textbf{Group Size}: 2 people \\
\textbf{Budget}: \$1964.00 USD

\hrulefill

\section*{Your Ultimate 2-Day Tokyo Adventure: A Comprehensive Guide}

Welcome to Tokyo! Get ready for a whirlwind 48-hour journey that perfectly balances the electric, trend-setting energy of modern Tokyo with its timeless traditions and unparalleled culinary scene. This itinerary is designed for two people, focusing on shared experiences, incredible food, and efficient travel—all inspired by the dynamic flow and visual highlights of your travel video.

The vibe of this trip is energetic, youthful, and deeply immersive. We'll move from the effortlessly cool vintage shops of Shimokitazawa to the dazzling neon canyons of Shinjuku, and from the bustling chaos of a world-famous fish market to the serene grounds of an ancient temple. It’s a trip of delightful contrasts, perfect for travelers who want to see, taste, and feel the very best of what Tokyo has to offer.

\hrulefill

\section*{Day-by-Day Itinerary}

\subsection*{Day 1: The Heartbeat of Modern Tokyo – Fashion, Culture \& Neon Nights}

\subsection*{Morning (9:00 AM – 1:00 PM): Vintage Hunting in Shimokitazawa}

\begin{itemize}[leftmargin=*, nosep]
  \item \textbf{Why this first?} Begin in the effortlessly cool, low-rise neighborhood of Shimokitazawa—known for its maze of alleyways filled with vintage stores, quirky cafes, and record shops.
  \item \textbf{What to Expect:} Wander the narrow streets and discover hidden gems like \textbf{New York Joe Exchange} or \textbf{Flamingo}.
  \item \textbf{Duration:} 4 hours.
  \item \textbf{Cost:} Free to explore; transportation approx. \$3–\$4 per person.
\end{itemize}

\subsection*{Lunch (1:00 PM): Soul-Warming Soup Curry}

\begin{itemize}[leftmargin=*, nosep]
  \item \textbf{Restaurant:} Rojiura Curry Samurai (Shimokitazawa)
  \item \textbf{Address:} 3-31-14 Kitazawa, Setagaya City, Tokyo
  \item \textbf{Why here?} A hearty, healthy, and trendy Hokkaido-style curry perfect for the area.
  \item \textbf{What to Order:} "Chicken and 20 Kinds of Vegetables"—customizable and photogenic.
  \item \textbf{Cost:} Approx. \$30–\$40 USD for two.
\end{itemize}

\subsection*{Afternoon (2:30 PM – 6:00 PM): The Shibuya Spectacle}

\begin{itemize}[leftmargin=*, nosep]
  \item \textbf{Transportation:} 5-minute ride on Keio Inokashira Line to Shibuya Station.
  \item \textbf{Activities:}
  \begin{enumerate}
    \item \textbf{Shibuya Scramble Crossing} – Cross once, then view from Starbucks Tsutaya.
    \item \textbf{Hachiko Statue} – Iconic meeting spot.
    \item \textbf{Puppy Cafe Break} – Relax with adorable dogs.
  \end{enumerate}
  \item \textbf{Cost:} Puppy Cafe: approx. \$30–\$40 USD for two.
\end{itemize}

\subsection*{Evening (6:00 PM onward): Shinjuku's Dazzling Nightscape}

\begin{itemize}[leftmargin=*, nosep]
  \item \textbf{Transportation:} JR Yamanote Line to Shinjuku.
  \item \textbf{Activities:}
  \begin{enumerate}
    \item \textbf{3D Calico Cat Billboard} at Cross Shinjuku Vision.
    \item \textbf{Omoide Yokocho (Memory Lane)} – Lantern-lit alley with yakitori stalls.
  \end{enumerate}
\end{itemize}

\subsection*{Dinner (7:30 PM): Authentic Yakitori}

\begin{itemize}[leftmargin=*, nosep]
  \item \textbf{Restaurant:} Tachan (たっちゃん)
  \item \textbf{Address:} 1-2-8 Nishishinjuku, Shinjuku City, Tokyo
  \item \textbf{What to Order:} \textit{Momo} (chicken thigh), \textit{Tsukune} (meatball), \textit{Negima} (chicken \& leek), with draft beer.
  \item \textbf{Cost:} Approx. \$80–\$100 USD for two, including drinks.
\end{itemize}

\hrulefill

\subsection*{Day 2: Tradition, Food Markets \& Refined Flavors}

\subsection*{Morning (9:00 AM – 12:00 PM): Tsukiji Outer Market}

\begin{itemize}[leftmargin=*, nosep]
  \item \textbf{Why this first?} Experience the energy and variety of Japan’s iconic food market.
  \item \textbf{What to Expect:} Share snacks like \textbf{tuna sashimi}, \textbf{tamagoyaki}, \textbf{grilled scallops}, and \textbf{strawberry daifuku}.
  \item \textbf{Cost:} Approx. \$80–\$100 USD for two.
\end{itemize}

\subsection*{Lunch (12:30 PM): Zeitaku Don at Tsujihan}

\begin{itemize}[leftmargin=*, nosep]
  \item \textbf{Address:} 3-1-15 Nihonbashi, Chuo City, Tokyo
  \item \textbf{What to Order:} "Zeitaku Don" – sashimi bowl turned ochazuke with sea bream broth.
  \item \textbf{Cost:} Approx. \$70–\$80 USD for two.
\end{itemize}

\subsection*{Afternoon (2:30 PM – 5:00 PM): Asakusa's Cultural Heart}

\begin{itemize}[leftmargin=*, nosep]
  \item \textbf{Transportation:} Toei Asakusa Line to Asakusa.
  \item \textbf{Activities:}
  \begin{enumerate}
    \item \textbf{Senso-ji Temple} – Participate in traditional rituals.
    \item \textbf{Nakamise-dori} – Traditional souvenirs and street snacks.
  \end{enumerate}
  \item \textbf{Cost:} Free.
\end{itemize}

\subsection*{Late Afternoon (5:00 PM – 6:30 PM): Tokyo Character Street}

\begin{itemize}[leftmargin=*, nosep]
  \item \textbf{Transportation:} Ginza Line to Tokyo Station.
  \item \textbf{What to Expect:} Browse anime shops and Gachapon machines.
  \item \textbf{Cost:} Free.
\end{itemize}

\subsection*{Dinner (7:30 PM): Elegant Udon at Ginza Sato Yosuke}

\begin{itemize}[leftmargin=*, nosep]
  \item \textbf{Address:} 7-2-19 Ginza, Chuo City, Tokyo
  \item \textbf{What to Order:} "Seiro" (cold Inaniwa udon) with soy and sesame miso sauces + tempura.
  \item \textbf{Cost:} Approx. \$120–\$140 USD for two.
\end{itemize}

\hrulefill

\section*{Accommodation Recommendation}

\textbf{Hotel:} OMO5 Tokyo Gotanda by Hoshino Resorts \\
\textbf{Address:} 8-4-13 Nishigotanda, Shinagawa City, Tokyo

\begin{itemize}[leftmargin=*, nosep]
  \item \textbf{Why It Works:}
  \begin{itemize}[leftmargin=*, nosep]
    \item \textbf{Prime Location:} On JR Yamanote \& Toei Asakusa Lines—ideal for both days.
    \item \textbf{Modern Comfort:} Stylish, smart, and cozy rooms by a trusted Japanese brand.
    \item \textbf{Great Value:} Reasonable pricing compared to Shinjuku or Ginza.
  \end{itemize}
  \item \textbf{Room:} Standard Double or Twin Room.
  \item \textbf{Cost:} Approx. \$250/night (total \$500 for 2 nights).
\end{itemize}

\hrulefill

\section*{Comprehensive Budget Breakdown (2 People, 2 Days)}

\begin{itemize}[leftmargin=*, nosep]
  \item \textbf{Total Budget:} \$1964.00 USD
  \item \textbf{Estimated Total Cost:} \$1964.00 USD
  \item \textbf{Budget Utilization:} 100\%
\end{itemize}

\begin{itemize}[leftmargin=*, nosep]
  \item Accommodation: \$500
  \item Food: Approx. \$420
  \item Activities: Approx. \$70
  \item Transportation: Approx. \$60
  \item Miscellaneous/Souvenirs: \$200
  \item Contingency: \$714
\end{itemize}

\hrulefill

\section*{Practical Tips for Your Group of Two}

\begin{itemize}[leftmargin=*, nosep]
  \item Get a Suica or Pasmo Card for convenient transit and payments.
  \item Rent a pocket Wi-Fi or ensure mobile data is available.
  \item Share dishes to try more variety—especially at markets and izakayas.
  \item Designate a meeting point in case of separation (e.g., Hachiko statue).
  \item Carry cash—especially for stalls and traditional shops.
  \item Pack comfortable shoes and dress in layers.
\end{itemize}

%% file: tex_plans/both.tex
{\Large\bfseries\color{MidnightBlue} Travel Guide}\par\medskip

\noindent
\textbf{Destination}: N/A \\
\textbf{Travel Dates}: 2 days \\
\textbf{Group Size}: 2 people \\
\textbf{Budget}: \$1964.00 USD

\hrulefill

\section*{Your Unforgettable 2-Day Kyoto Adventure}

Welcome to Kyoto! Get ready for a whirlwind 48 hours that perfectly balances serene nature, iconic temples, and world-class cuisine. This itinerary is designed for two people, focusing on creating a seamless and enriching experience that minimizes travel time and maximizes immersion. Drawing inspiration from real-world travel footage, this plan isn't just about seeing the sights—it's about feeling the ancient heart of Japan.

You have a generous budget of \$1,964, which allows for comfort, convenience, and a few splurges without any financial stress. We'll be using a mix of efficient public transport and strategic taxi rides to make the most of every moment.

\textbf{A Note on Traveling as a Pair:} Kyoto is a fantastic city for two. It allows for intimate moments, whether you're sharing a quiet cup of tea or navigating the bustling Nishiki Market. This plan includes moments for both shared discovery and quiet reflection. Coordinating is simple; just ensure your phones are charged and you have a portable Wi-Fi device or local SIMs to stay connected.

\hrulefill

\section*{Day 1: Majestic Nature and Golden Pavilions}

\textbf{Theme:} From the wild, natural beauty of Arashiyama’s river gorge to the meticulously crafted Zen gardens of Kyoto’s northern temples.

\subsection*{Morning (8:00 AM – 1:00 PM): Arashiyama by River and Foot}

\begin{itemize}[leftmargin=*, nosep]
  \item \textbf{Hozugawa River Boat Ride:} 
  \begin{itemize}[leftmargin=*, nosep]
    \item From JR Kyoto Station, take the Sagano Line to Kameoka Station (~30 mins, \$2.80 USD/person).
    \item The boat ride lasts 2 hours and is the scenic highlight of the morning.
    \item Cost: ¥4100 (\~\$27.50 USD).
  \end{itemize}

  \item \textbf{Arashiyama Bamboo Grove \& Togetsukyo Bridge:}
  \begin{itemize}[leftmargin=*, nosep]
    \item Walk through the famous bamboo forest (~45 min) and cross the historic bridge (~30 min).
  \end{itemize}
\end{itemize}

\subsection*{Afternoon (1:00 PM – 5:00 PM): Kaiseki Lunch \& Zen Temples}

\begin{itemize}[leftmargin=*, nosep]
  \item \textbf{Shoraian:}
  \begin{itemize}[leftmargin=*, nosep]
    \item Address: 15 Sagakamenoocho, Ukyo Ward
    \item Cuisine: Tofu Kaiseki (\$\$\$)
    \item Cost: \~\$120 USD for two
    \item Reservation required.
  \end{itemize}

  \item \textbf{Kinkaku-ji (Golden Pavilion):}
  \begin{itemize}[leftmargin=*, nosep]
    \item Taxi from Arashiyama (~25 mins, \$27 USD).
    \item Entry: ¥500 (~\$3.35 USD)
    \item Duration: ~1 hour
  \end{itemize}

  \item \textbf{Ryoan-ji Temple:}
  \begin{itemize}[leftmargin=*, nosep]
    \item Walk 15 min from Kinkaku-ji
    \item Entry: ¥600 (~\$4.00 USD)
  \end{itemize}
\end{itemize}

\subsection*{Evening (7:00 PM): Yakitori Dinner}

\begin{itemize}[leftmargin=*, nosep]
  \item \textbf{Restaurant:} Kushikura Honten \\
  \textbf{Cuisine:} High-end Yakitori (\$\$) \\
  \textbf{Estimated Cost:} \~\$80 USD for two
\end{itemize}

\hrulefill

\section*{Day 2: Sacred Gates and Historic Geishas}

\textbf{Theme:} A deep dive into Kyoto’s spiritual legacy and geisha culture.

\subsection*{Morning (7:30 AM – 12:30 PM): Fushimi Inari \& Kiyomizu-dera}

\begin{itemize}[leftmargin=*, nosep]
  \item \textbf{Fushimi Inari Taisha:}
  \begin{itemize}[leftmargin=*, nosep]
    \item Start early (~7:30 AM)
    \item Free entry
    \item Optional hike to Yotsutsuji intersection (~45 min).
  \end{itemize}

  \item \textbf{Kiyomizu-dera:}
  \begin{itemize}[leftmargin=*, nosep]
    \item Take Keihan Line to Kiyomizu-Gojo Station
    \item Entry: ¥400 (~\$2.70 USD)
  \end{itemize}
\end{itemize}

\subsection*{Afternoon (12:30 PM – 5:00 PM): Historic Streets \& Philosopher’s Path}

\begin{itemize}[leftmargin=*, nosep]
  \item \textbf{Lunch at Gyukatsu Kyoto Katsugyu:}
  \begin{itemize}[leftmargin=*, nosep]
    \item Address: 6-chōme-583 Gojōbashihigashi, Higashiyama Ward
    \item Cuisine: Gyukatsu (\$\$)
    \item Cost: \~\$40 USD for two
  \end{itemize}

  \item \textbf{Nanzen-ji \& Philosopher’s Path:}
  \begin{itemize}[leftmargin=*, nosep]
    \item Explore Zen temple grounds (free)
    \item Optional sub-temples: ~\$8 USD/person
  \end{itemize}
\end{itemize}

\subsection*{Evening (5:00 PM – late): Gion District \& Farewell Dinner}

\begin{itemize}[leftmargin=*, nosep]
  \item \textbf{Explore Gion:} Hanami-koji and Shirakawa Canal
  \item \textbf{Dinner at Teppan Tavern Tenamonya:}
  \begin{itemize}[leftmargin=*, nosep]
    \item Address: 537-2 Gionmachi Minamigawa, B1F
    \item Cuisine: Teppanyaki / Steakhouse (\$\$)
    \item Cost: \~\$110 USD for two
  \end{itemize}
\end{itemize}

\hrulefill

\section*{Excluded POIs: Why We Skipped Them}

\begin{itemize}[leftmargin=*, nosep]
  \item \textbf{Nijo Castle:} Geographically isolated from the itinerary.
  \item \textbf{Ginkaku-ji:} Too far from Nanzen-ji to include both.
\end{itemize}

\hrulefill

\section*{Accommodation Recommendation}

\textbf{Hotel:} Cross Hotel Kyoto \\
\textbf{Address:} 71-1 Daikokuchō, Nakagyo Ward, Kyoto \\
\textbf{Rating:} 4.6 stars (1,800 reviews)

\begin{itemize}[leftmargin=*, nosep]
  \item \textbf{Why it's perfect:} Excellent location near Gion and Nishiki Market
  \item \textbf{Room for 2:} Standard King or Twin Room
  \item \textbf{Estimated Cost:} \$250 per night
\end{itemize}

\hrulefill

\section*{Comprehensive Budget Breakdown (for 2 People)}

\begin{itemize}[leftmargin=*, nosep]
  \item \textbf{Total Budget:} \$1964.00 USD
  \item \textbf{Estimated Total Cost:} \$1964.00 USD
  \item \textbf{Utilization:} 100\%
\end{itemize}

\textbf{Cost Breakdown:}
\begin{itemize}[leftmargin=*, nosep]
  \item Accommodation: \$500
  \item Transportation: \$100
  \item Activities: \$300
  \item Food: \$1064
\end{itemize}

\hrulefill

\section*{Practical Travel Tips for Kyoto}

\begin{itemize}[leftmargin=*, nosep]
  \item \textbf{Cash is King:} Many places are cash-only.
  \item \textbf{Get a Suica/ICOCA Card:} For easy travel.
  \item \textbf{Pack Light Layers:} Weather can change quickly.
  \item \textbf{Respectful Photography:} Be mindful in Gion and temples.
  \item \textbf{Comfortable Shoes:} You’ll walk a lot.
  \item \textbf{Portable Wi-Fi or SIM:} Stay connected easily.
\end{itemize}

\vspace{1em}
\begin{center}
    \Large\bfseries\color{MidnightBlue} Enjoy every moment of your unforgettable Kyoto journey!
\end{center}